\documentclass[lettersize,journal]{IEEEtran}
\usepackage{amsmath,amsfonts}
\usepackage[ruled,vlined]{algorithm2e}
\usepackage{array}
\usepackage{colortbl}
\usepackage[caption=false,font=normalsize,labelfont=sf,textfont=sf]{subfig}
\usepackage{textcomp}
\usepackage{stfloats}
\usepackage{url}
\usepackage{verbatim}
\usepackage{graphicx}
\usepackage{cite}
\usepackage{tabulary}
\usepackage[dvipsnames]{xcolor}
\usepackage{booktabs}
\usepackage{multirow}
\usepackage{cuted}
\usepackage{caption}
\usepackage[breaklinks,colorlinks]{hyperref}
\newcommand{\tablestyle}[2]{\setlength{\tabcolsep}{#1}\renewcommand{\arraystretch}{#2}\centering\footnotesize}
\begin{document}

\title{Generalized Small Object Detection: \\A Point-Prompted Paradigm and Benchmark}
\author{Haoran Zhu, 
Wen Yang,~\IEEEmembership{Senior Member,~IEEE},
Guangyou Yang, 
Chang Xu, 
Ruixiang Zhang, \\
Fang Xu, 
Haijian Zhang,~\IEEEmembership{Senior Member,~IEEE}, 
Gui-Song Xia,~\IEEEmembership{Senior Member,~IEEE}
\thanks{H Zhu, W. Yang, G. Yang, R. Zhang, and H. Zhang are with the School of Electronic Information, Wuhan University, Wuhan, 430072 China. \emph{E-mail: \{zhuhaoran, yangwen, yangguangyou, zhangruixiang, haijian.zhang\}@whu.edu.cn}}
\thanks{C. Xu is with the Environmental Computational Science and Earth Observation Laboratory, EPFL, Sion, Switzerland. \emph{E-mail: chang.xu@epfl.ch}}
\thanks{F. Xu and G-S. Xia are with the School of Artificial Intelligence, Wuhan University, Wuhan, 430072, China. \emph{E-mail: xufang, guisong.xia@whu.edu.cn}}
}

\markboth{Journal of \LaTeX\ Class Files,~Vol.~14, No.~8, August~2021}%
{Shell \MakeLowercase{\textit{et al.}}: A Sample Article Using IEEEtran.cls for IEEE Journals}

\maketitle

\begin{strip}
    \centering
    \vspace{-90pt}
    \includegraphics[width=\linewidth]{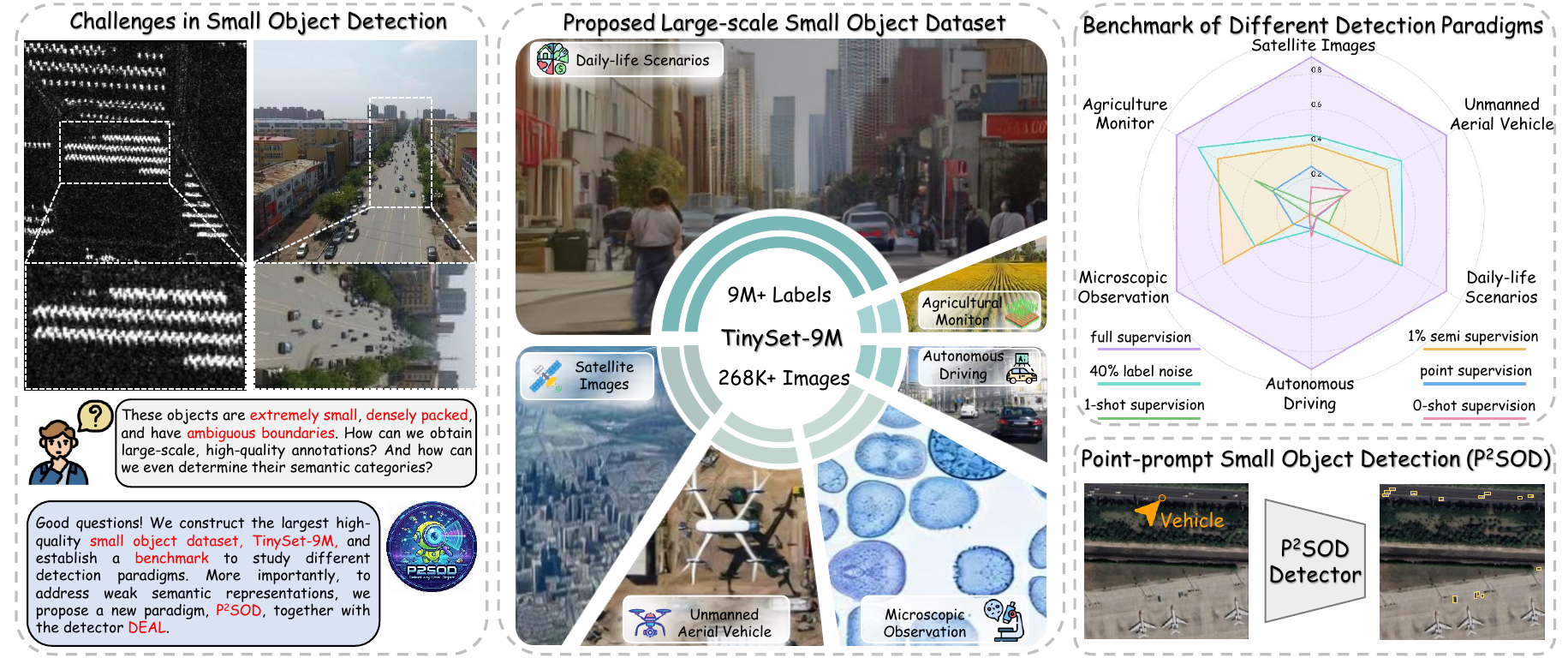}
    \captionof{figure}{Overview of our study on generalized small object detection. Leveraging the proposed TinySet-9M dataset and benchmark, we systematically investigate the performance of existing label-efficient paradigms in the small-object regime and introduce a new detection paradigm, Point-prompt Small Object Detection (P$^2$SOD). The middle panel illustrates the domain composition of TinySet-9M, while the right panel compares the performance of representative label-efficient paradigms on small objects and our proposed detection paradigm.}
    \label{fig:first_figure}
    \vspace{0pt}
\end{strip}

\begin{abstract}
Small object detection (SOD) remains challenging due to extremely limited pixels and ambiguous object boundaries. These characteristics lead to challenging annotation, limited availability of large-scale high-quality datasets, and inherently weak semantic representations for small objects.
In this work, we first address the data limitation by introducing TinySet-9M, the first large-scale, multi-domain dataset for small object detection. Beyond filling the gap in large-scale datasets, we establish a benchmark to evaluate the effectiveness of existing label-efficient detection methods for small objects. Our evaluation reveals that weak visual cues further exacerbate the performance degradation of label-efficient methods in small object detection, highlighting a critical challenge in label-efficient SOD.
Secondly, to tackle the limitation of insufficient semantic representation, we move beyond training-time feature enhancement and propose a new paradigm termed Point-Prompt Small Object Detection (P$^2$SOD). This paradigm introduces sparse point prompts at inference time as an efficient information bridge for category-level localization, enabling semantic augmentation.
Building upon the P$^2$SOD paradigm and the large-scale TinySet-9M dataset, we further develop DEAL (DEtect Any smalL object), a scalable and transferable point-prompted detection framework that learns robust, prompt-conditioned representations from large-scale data. 
With only a single click at inference time, DEAL achieves a 31.4\% relative improvement over fully supervised baselines under strict localization metrics (\textit{e.g.}, AP75) on TinySet-9M, while generalizing effectively to unseen categories and unseen datasets.
Our project is available at \href{https://zhuhaoraneis.github.io/TinySet-9M/}{GitHub}.
\end{abstract}


\section{Introduction}
\label{sec:intro}
\IEEEPARstart{D}{etecting} small objects is a long-standing challenge in object detection, arising in a wide range of applications such as remote sensing, autonomous driving, and microscopic analysis~\cite{DCFL++, soda_2023_pami}.
Despite continuous advances in detection architectures and training strategies, the performance gap between small objects and generic objects remains persistent across datasets and paradigms.
This difficulty primarily stems from the intrinsic properties of small objects. Typically occupying only a few pixels (\textit{e.g.}, smaller than $32\times32$ pixels~\cite{COCO_2014_ECCV}), small objects exhibit blurred boundaries, weak semantic cues, and severe inter-object ambiguity, especially in dense scenes. 
These properties lead to two direct and practical consequences.
On the one hand, minor spatial deviations can cause drastic drops in IoU-based metrics~\cite{aitodv2_2022_isprs}, while dense layouts and ambiguous boundaries make precise annotation highly error-prone and often infeasible at scale~\cite{labelnoise_dntod}. As a result, constructing large-scale, high-quality datasets for small object detection requires substantial human effort and cost.
On the other hand, small objects provide insufficient visual evidence for reliable localization and category recognition. Weak semantic cues make category discrimination highly unstable, especially in cluttered scenes, which fundamentally limits the accuracy and generalization ability of small object detectors.
Consequently, the SOD community has been constrained by two long-standing challenges that this work aims to address:
(1) \textbf{the scarcity of large-scale, high-quality datasets} due to prohibitive annotation costs, and
(2) \textbf{the difficulty of achieving accurate and generalized small object detection} under weak semantic signals.

\begin{figure*}[t]
\centering
\includegraphics[width=\linewidth]{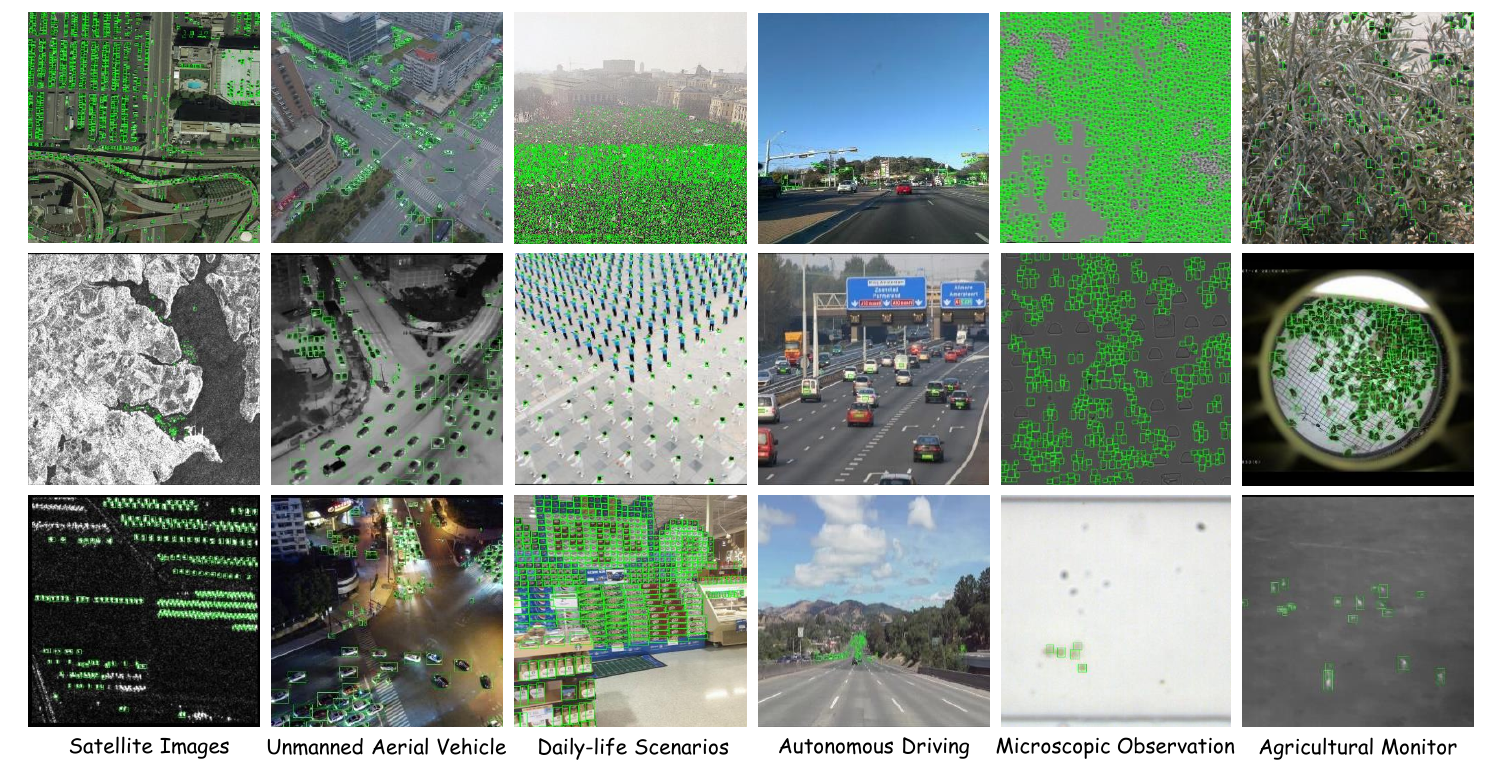}
\caption{Visualization of the TinySet-9M dataset. The dataset contains a large number of densely distributed objects with extremely small spatial scales, which results in blurred object boundaries and weak semantic representations.}
\label{fig:tinyset_9m_visualization}
\end{figure*}

To address the first challenge, we focus on the scalability of data and supervision.
Existing small object datasets are typically limited in data scale~\cite{TinyPerson_2020_WACV, aitodv2_2022_isprs}, domain coverage~\cite{rgbt-tiny, RGBTDronePerson}, or annotation diversity~\cite{soda_2023_pami, visdrone_2021_pami}, making it difficult to advance the practical application and generalization of small object detectors.
In this work, we introduce TinySet-9M, a large-scale, multi-domain dataset for small object detection, containing over 9 million annotations across six representative domains (as illustrated in Fig.~\ref{fig:first_figure} and Fig.~\ref{fig:tinyset_9m_visualization}). 
Using this dataset, we benchmark a wide range of supervision regimes, including noisy annotations, semi-supervised learning, point supervision, sparse-shot learning, and zero-shot detection.
Our evaluation reveals a conclusion: weak visual cues of small objects further exacerbate the performance degradation of label-efficient methods in SOD. 
As shown in Table~\ref{table:bench_tinyset_10m} and Fig.~\ref{fig:tinyset_9m_benchmark_detection}, under 40\% label noise, small objects (TinySet-9M) achieve only 26.9\% of fully supervised performance, whereas generic objects (COCO) retain 49.7\%~\cite{labelnoise_oamil}.
Under the 10\% labeled semi-supervised setting, small objects achieve 69.2\% of fully supervised performance, while generic objects reach 78.1\%~\cite{pseco}.
Label-efficient methods that perform well in generic object detection fail to maintain reliable localization and recognition under SOD task, indicating that small objects require dedicated designs that can reduce annotation costs while preserving detection accuracy.

The second challenge lies in the representation limitation stemming from the inherent semantic weakness of small objects.
Existing methods attempt to improve small object detection through training-time feature enhancement~\cite{tod_hsfpn, related_SET} or optimization strategies~\cite{tod_ga, tod_simd}.
However, due to their extremely limited pixel footprint, the features of small objects tend to vanish during the progressive downsampling process of deep networks (\textit{e.g.}, a $32 \times 32$ object reduces to only $4 \times 4$ feature points after an $8\times$ downsampling). Consequently, it is inherently difficult to sufficiently amplify or preserve their representations. Therefore, we propose a shift in perspective: instead of relying solely on training-time feature enhancement, we explore whether minimal and targeted information introduced at inference time can compensate for the semantic deficiency inherent to small objects.

Based on this insight, we propose a new paradigm termed Point-Prompt Small Object Detection (P$^2$SOD). P$^2$SOD introduces sparse point prompts at inference time as an efficient information bridge for category-level localization. A point prompt provides precise spatial grounding with low interaction cost, making it particularly suitable for small objects whose visual boundaries and semantics are ambiguous. Beyond merely supplementing weak visual features, acquiring such point priors seamlessly addresses critical bottlenecks in real-world pipelines. For instance, a single expert click in human-in-the-loop systems can instantly retrieve all target instances across gigapixel imagery~\cite{irtdetr}, while coarse coordinates from sensors (\textit{e.g.}, radar) can naturally serve as zero-cost point priors to guide detection in automated systems (Further discussions, detailed analysis, and comparisons with SAM-based methods are provided in Section~\ref{sec:discussion}).
Building upon the P$^2$SOD paradigm and the large-scale TinySet-9M dataset, we further develop DEAL (DEtect Any smalL object), a scalable and transferable point-prompted detection framework. 
DEAL extends conventional image-only detectors to support point-conditioned reasoning, enabling robust and transferable small object detection under sparse interaction.
Specifically, DEAL departs from standard detection pipelines by explicitly modeling the interaction between point prompts and visual features, and by allowing the supervision signals to be dynamically conditioned on the input point prompts during training. It incorporates a Point-Guided Density Activation mechanism to inject category-level point information into multi-scale feature representations, selectively amplifying responses of objects belonging to the prompted category. Meanwhile, a Prediction-Guided Cyclic Prompting strategy is employed to progressively refine predictions under varying prompt configurations, allowing the detector to learn stable and generalizable point-conditioned representations.
Under this paradigm, our DEAL achieves efficient and accurate small object detection under sparse point prompts without increasing memory consumption. Notably, with only a single point prompt per category, our method surpasses the fully supervised RT-DETR by 12.1 AP points. More importantly, it demonstrates strong generalization to unseen datasets and categories, achieving 55.9 AP on DOTA-v2.0 and significantly outperforming state-of-the-art VLM-based methods such as SAM3 and Rex-Omni.

Our main contributions are summarized as follows:

\begin{itemize}
    \item We introduce TinySet-9M, the first large-scale, multi-domain dataset designed for small object detection, containing over 9 million annotations across six domains.
    \item We establish a benchmark on TinySet-9M, and observe a consistent performance collapse under reduced supervision, identifying localization instability caused by insufficient visual and semantic evidence as an important limitation of small object detection.
    \item We propose Point-Prompt Small Object Detection (P$^2$SOD) and develop DEAL, a scalable point-prompted detector that leverages minimal inference-time interaction to achieve accurate and generalized small object detection across unseen categories and datasets.
\end{itemize}

\begin{figure*}[t]
\centering
\includegraphics[width=\linewidth]{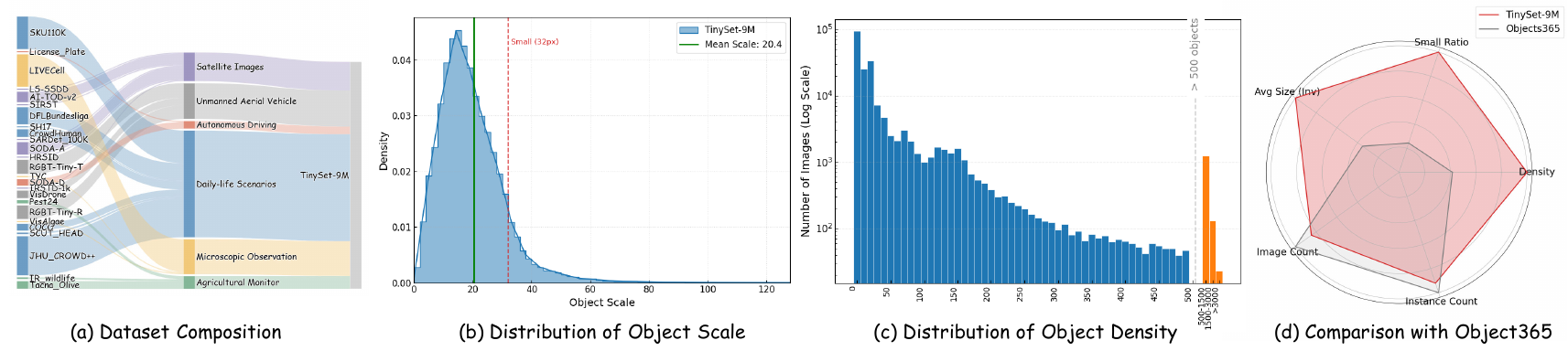}
\caption{Statistical analysis of the TinySet-9M dataset. (a) shows the composition of TinySet-9M, where different source datasets form distinct sub-domains; (b) presents the distribution of object scales, in which the green line indicates the average object scale of the dataset (20.4) and the red line denotes the scale threshold for small objects; (c) illustrates the distribution of object density across images; (d) shows the comparison of TinySet-9M and Object365 in terms of the proportion of small objects, object density, average object scale, number of instances, and number of images.}
\label{fig:tinyset_9m_statics}
\end{figure*}

\section{Related works}
\subsection{Small Object Detection}
Small object detection differs fundamentally from generic object detection due to the limited spatial extent, ambiguous boundaries, and frequent dense distributions of small objects. These characteristics not only increase the difficulty of reliable localization and recognition, but also make accurate annotation labor-intensive and error-prone, posing challenges for both algorithm design and benchmark construction.

Early studies on small object detection typically relied on extracting small-object subsets from generic datasets such as COCO~\cite{COCO_2014_ECCV} and VOC~\cite{PASCAL-VOC_2015-IJCV}. With the growing demand from real-world applications, several domain-specific benchmarks have been introduced in recent years, particularly in remote sensing, UAV imagery, autonomous driving, and biomedical imaging. Representative datasets include AI-TOD-v2~\cite{aitodv2_2022_isprs}, TinyPerson~\cite{TinyPerson_2020_WACV}, RGBTDronePerson~\cite{RGBTDronePerson}, SODA~\cite{soda_2023_pami}, and LiveCell~\cite{livecell}, each focusing on small object detection within a specific domain or sensing modality. While these benchmarks have significantly advanced domain-level performance, most of them remain limited in scale and domain diversity, and predominantly assume fully supervised annotations, making it difficult to assess detection performance under realistic annotation budgets.

From an algorithmic perspective, existing small object detection methods primarily aim to improve training-time learning under full supervision. Representative approaches refine optimization and supervision mechanisms to address the limitations of IoU-based metrics at small scales, including alternative similarity measures and improved label assignment strategies (\textit{e.g.}, NWD~\cite{aitodv2_2022_isprs}, SimD~\cite{related_SimD_strategy}, RFLA~\cite{rfla_2022_eccv}). Other methods enhance the representational capacity of small objects through architectural design, such as attention-based feature modulation, feature reconstruction, multi-receptive-field modeling, and information-driven feature amplification (\textit{e.g.}, FRLI-Net~\cite{related_FRLI_Net}, PFIM~\cite{related_PFIM}).
Despite their effectiveness, these approaches generally rely on dense and high-quality annotations, which limit their scalability when annotation cost becomes a primary concern.

In summary, existing datasets and methods have driven steady progress in small object detection, yet current benchmarks remain constrained in scale and supervision diversity, while most algorithms depend heavily on fully supervised training and image-only semantics. These limitations motivate our large-scale benchmark construction and the exploration of point-prompted detection as a scalable solution for generalized small object detection.

\subsection{Label-efficient Object Detection}
Label-efficient learning has emerged as a crucial paradigm for training object detectors under limited annotation budgets, playing an essential role in the construction of high-cost object detection datasets. According to the form of low-cost supervision, existing label-efficient object detection approaches can be broadly categorized into label-noise–robust learning, semi-supervised learning, weakly supervised learning, sparse-shot learning, and zero-shot learning.
These paradigms share the common objective of reducing annotation cost, yet differ substantially in their underlying assumptions and learning mechanisms:

Label-noise–robust learning focuses on improving robustness under imperfect annotations, such as bounding-box shifts, missing labels, or category noise, often adopting Multiple Instance Learning (MIL) frameworks to correct localization and classification errors~\cite{labelnoise_ssddet, labelnoise_oamil, labelnoise_dntod}. Semi-supervised learning combines labeled and unlabeled data, typically using teacher–student frameworks with pseudo-label generation and refinement to progressively improve detection accuracy on unlabeled images~\cite{semi_softteacher, pseco, semi_consistent_teacher}. Weakly supervised learning further reduces annotation granularity by relying on coarse supervision, such as point-level, image-level, or scribble annotations, and commonly converts them into bounding boxes through proposal generation and MIL-based optimization~\cite{p2bnet, point_teacher, point2rboxv3, pointobbv3}. Sparse-shot learning addresses scenarios with only a few annotated instances per image, leveraging teacher–student or Siamese architectures to propagate limited supervision and discover unlabeled objects~\cite{sparse_calibratedteacher, ALOD, sparse_comining, sparse_sparsedet}. More recently, zero-shot detection has emerged with large-scale foundation models, including open-vocabulary object detection methods that align visual regions with text embeddings~\cite{groundingdino, groundedsam, VLM_CLIP}, as well as visual-prompt-based approaches that use minimal visual cues, such as points or boxes, as semantic anchors for generic object detection~\cite{sam3, rexomni}.

Although these approaches have been extensively validated and widely adopted in generic object detection, their effectiveness on small object detection has not yet been systematically analyzed. In parallel, it remains unclear whether these algorithms exhibit consistent performance gains as data scale increases. To address these problems, we select two to three representative methods from each label-efficient paradigm and evaluate their performance on the large-scale small object detection benchmark TinySet-9M.

\section{TinySet-9M Dataset}
\label{sec:dataset}

We identify six representative domains in which small objects are most commonly observed, including Satellite Images, Unmanned Aerial Vehicles, Daily-life Scenarios, Autonomous Driving, Microscopic Observation, and Agricultural Monitoring. Based on this domain taxonomy, we conduct a systematic survey of high-quality small object datasets and ultimately select 23 representative datasets for integration.
For the Satellite Images domain, the collected datasets include AI-TOD-v2.0~\cite{aitodv2_2022_isprs}, SODA-A~\cite{soda_2023_pami}, HRSID~\cite{hrsid}, IRSTD-1K~\cite{irstd-1k}, LS-SSDD~\cite{lsssdd}, SARDet-100K~\cite{sardet100k}, and SIRST~\cite{sirst}.
For the Unmanned Aerial Vehicles domain, we incorporate RGBT-Tiny~\cite{rgbt-tiny} and VisDrone~\cite{visdrone_2021_pami}.
For Daily-life Scenarios, we include widely used benchmarks such as COCO~\cite{COCO_2014_ECCV}, CrowdHuman~\cite{crowdhuman}, DFL-Bundesliga~\cite{DFL-BundesligaDataset}, JHU\_CROWD++~\cite{jhu_crowd++}, SCUT\_Head~\cite{SCUTHEADDataset}, SH17~\cite{SH17Dataset}, and SKU110K~\cite{sku110k}.
For the Autonomous Driving domain, the selected datasets consist of SODA-D~\cite{soda_2023_pami} and License\_Plate~\cite{LicensePlateDetectionDataset}.
For Microscopic Observation, we collect LIVECell~\cite{livecell}, TYC~\cite{tyc}, and VisAlgae~\cite{visalgae}, which capture small objects at cellular and microorganism scales.
For Agricultural Monitoring, we incorporate IR\_Wildlife~\cite{IRwildlifeDataset}, Pest24~\cite{pest24}, and Tacna\_Olive~\cite{tacna_olive}, covering small object detection tasks in natural and agricultural environments.
The relative proportions of datasets from different domains are illustrated in Fig.~\ref{fig:tinyset_9m_statics}(a).
To further refine the dataset, we filter out images in which large objects occupy more than 40\% of the image area, ensuring a consistent focus on small object detection. In addition, all images are either cropped or resized with padding to a unified resolution of $1024 \times 1024$. 
We summarize the characteristics of this dataset with the following aspects.

\textbf{Domain composition}.
Fig.~\ref{fig:tinyset_9m_statics}(a) presents the domain-wise composition of TinySet-9M. The Daily-life Scenarios domain constitutes the largest portion of the dataset, covering commonly occurring objects such as shelf items and pedestrians. The Satellite Images domain contributes a wide range of sensing modalities, including visible, infrared, and SAR imagery, reflecting the multi-modal nature of small object detection in aerial and satellite scenarios.

\textbf{Object scale distribution}.
Fig.~\ref{fig:tinyset_9m_statics}(b) illustrates the distribution of object scales in TinySet-9M. A substantial fraction of instances fall within the 0–20 pixel range, and the average object scale is 20.4 pixels. For reference, this average scale is notably smaller than those reported for large-scale natural image datasets such as COCO, Object365, and SA-1B, indicating that TinySet-9M places greater emphasis on small-scale object instances.

\textbf{Object density distribution}.
Fig.~\ref{fig:tinyset_9m_statics}(c) shows the distribution of object density across images. Many images contain more than 200 annotated objects, and a small subset exceeds 3,000 instances per image. This distribution reflects the prevalence of crowded scenes in several constituent domains, such as satellite images and daily-life surveillance.

\textbf{Comparison with large-scale benchmarks}.
Fig.~\ref{fig:tinyset_9m_statics}(d) provides a quantitative comparison between TinySet-9M and Object365 in terms of dataset scale and object statistics. While TinySet-9M contains fewer images and total instances, it exhibits a higher proportion of small objects, smaller average object scales, and higher object densities. This comparison highlights the complementary statistical properties of TinySet-9M relative to general-purpose detection benchmarks.
\section{TinySet-9M Benchmark}
\label{sec:benchmark}

\setlength{\tabcolsep}{3pt}
\begin{table*}[t]
\centering
\caption{Main results of fully-supervised, noise-supervised, semi-supervised, sparse-annotated, point-supervised, sparse-shot, and zero-shot methods on TinySet-9M (class-agnostic). For the training schedule, 1$\times$ denotes 3 epochs. All experiments are run on a computer with an NVIDIA RTX 3090 (24 GB) GPU. We use FP32 with 1024 $\times$ 1024 inputs.}
\label{table:bench_tinyset_10m}
\renewcommand{\arraystretch}{0.6}
\renewcommand{\tabcolsep}{1.0mm}
\resizebox{0.9\linewidth}{!}{
\begin{tabular}{l|l|c|c|c|ccc|cccc}
	\toprule
	ID & Method & Setting & Backbone & Schedule & $\rm{AP_{0.25}}$ & $ \rm{AP_{0.5}}$ & $\rm{AP_{0.75}}$ & $\rm{AP_{vt}}$ & $\rm{AP_{t}}$ & $\rm{AP_{s}}$ & $\rm{AP_{m}}$ \\
	\midrule
    \multicolumn{12}{l}{\textit{fully-supervised:}} \\
    \midrule
    \#1 & RetinaNet~\cite{Focal-Loss_2017_ICCV} & \multirow{6}{*}{100\% labeled} & ResNet-50 & 1$\times$ & 56.7 & 37.8 & 10.6 & 6.5 & 25.0 & 49.7 & 83.3 \\
    \#2 & Faster R-CNN~\cite{Faster-R-CNN_2015_NIPS} &  & ResNet-50 & 1$\times$ & 51.9 & 43.2 & 17.8 & 5.0 & 30.5 & 59.0 & 84.5 \\
    \#3 & FCOS~\cite{FCOS_2022_TPAMI} &  & ResNet-50 & 1$\times$ & 66.3 & 49.5 & 17.1 & 13.8 & 41.1 & 62.3 & 83.7 \\
    \#4 & RT-DETR~\cite{rtdetr} &  & ResNet-50 & 1$\times$ & 57.2 & 43.1 & 12.1 & 11.3 & 35.3 & 54.7 & 74.1 \\
    \#5 & HS-FPN (Faster R-CNN)~\cite{tod_hsfpn} & & ResNet-50 & 1$\times$ & 51.8 & 43.0 & 17.5 & 5.9 & 31.2 & 59.1 & 83.0 \\ 
    \#6 & RFLA (Faster R-CNN)~\cite{rfla_2022_eccv} &  & ResNet-50 & 1$\times$ & 66.9 & 54.6 & 20.8 & 15.3 & 46.6 & 67.5 & 88.4 \\ 
    \midrule
    \midrule
    \multicolumn{12}{l}{\textit{label noise:}} \\
    \midrule
    \#7 & FCOS & \multirow{4}{*}{20\% box noise} & ResNet-50 & 1$\times$ & 66.8 & 47.7 & 12.5 & 12.4 & 40.6 & 59.5 & 81.9 \\   
    \#8 & Faster R-CNN & & ResNet-50 & 1$\times$ & 45.2 & 33.8 & 8.1 & 4.0 & 26.3 & 44.7 & 73.1 \\
    \#9 & DN-TOD (FCOS)~\cite{labelnoise_dntod} & & ResNet-50 & 1$\times$ & 66.9 & 47.1 & 13.2 & 13.1 & 38.7 & 59.3 & 81.6 \\
    \#10 & SSD-Det (Faster R-CNN)~\cite{labelnoise_ssddet} & & ResNet-50 & 1$\times$ & 49.0 & 37.2 & 10.0 & 4.9 & 27.2 & 49.9 & 73.5 \\
    \midrule
    \#11 & FCOS & \multirow{4}{*}{40\% box noise} & ResNet-50 & 1$\times$ & 52.8 & 33.5 & 2.7 & 9.5 & 30.7 & 43.2 & 55.8 \\   
    \#12 & Faster R-CNN & & ResNet-50 & 1$\times$ & 23.2 & 11.6 & 0.3 & 2.0 & 21.7 & 14.3 & 25.5 \\
    \#13 & DN-TOD (FCOS)~\cite{labelnoise_dntod} & & ResNet-50 & 1$\times$ & 60.2 & 38.2 & 4.6 & 10.1 & 29.9 & 48.5 & 71.4 \\
    \#14 & SSD-Det (Faster R-CNN)~\cite{labelnoise_ssddet} & & ResNet-50 & 1$\times$ & 47.5 & 32.1 & 4.9 & 5.0 & 23.6 & 41.5 & 68.5 \\
    \midrule
    \midrule
    \multicolumn{12}{l}{\textit{semi-supervised:}} \\
    \midrule
    \#15 & Faster R-CNN & \multirow{3}{*}{1\% labeled} & ResNet-50 & 1$\times$ & 36.8 & 25.7 & 7.1 & 2.9 & 20.0 & 33.7 & 55.8 \\
    \#16 & Soft Teacher~\cite{semi_softteacher} & & ResNet-50 & 1$\times$ & 38.9 & 27.9 & 7.3 & 2.0 & 23.3 & 36.9 & 54.1 \\
    \#17 & Pseco~\cite{pseco} & & ResNet-50 & 1$\times$ & 43.2 & 30.6 & 9.0 & 3.0 & 26.1 & 40.2 & 59.5 \\
    \midrule
    \#18 & Faster R-CNN & \multirow{3}{*}{5\% labeled} & ResNet-50 & 1$\times$ & 41.5 & 29.3 & 8.5 & 3.0 & 24.1 & 39.1 & 63.1 \\
    \#19 & Soft Teacher~\cite{semi_softteacher} & & ResNet-50 & 1$\times$ & 41.6 & 30.6 & 8.6 & 3.0 & 24.9 & 40.4 & 62.5 \\
    \#20 & Pseco~\cite{pseco} & & ResNet-50 & 1$\times$ & 45.6 & 32.8 & 10.1 & 4.0 & 27.0 & 43.6 & 63.5 \\
    \midrule
    \#21 & Faster R-CNN & \multirow{3}{*}{10\% labeled} & ResNet-50 & 1$\times$ & 42.2 & 29.9 & 8.6 & 2.9 & 24.3 & 40.0 & 64.1 \\
    \#22 & Soft Teacher~\cite{semi_softteacher} & & ResNet-50 & 1$\times$ & 39.8 & 28.1 & 7.3 & 3.0 & 23.1 & 36.6 & 58.9 \\
    \#23 & Pseco~\cite{pseco} & & ResNet-50 & 1$\times$ & 46.5 & 33.4 & 10.1 & 4.0 & 27.8 & 43.9 & 64.7 \\
    \midrule
    \midrule
    \multicolumn{12}{l}{\textit{sparse-shot:}} \\
    \midrule
    \#24 & Faster R-CNN & \multirow{3}{*}{1-shot} & ResNet-50 & 1$\times$ & 7.4 & 5.1 & 1.1 & 2.0 & 4.5 & 8.6 & 3.3 \\
    \#25 & Co-mining~\cite{sparse_comining} & & ResNet-50 & 1$\times$ & 7.9 & 5.9 & 1.6 & 2.0 & 5.2 & 9.6 & 3.9 \\
    \#26 & Calibrated Teacher~\cite{sparse_calibratedteacher} & & ResNet-50 & 1$\times$ & 7.2 & 5.1 & 1.3 & 2.0 & 4.4 & 8.6 & 3.1 \\
    \midrule
    \#27 & Faster R-CNN & \multirow{3}{*}{3-shot} & ResNet-50 & 1$\times$ & 14.5 & 10.0 & 2.6 & 2.0 & 8.7 & 14.9 & 13.8 \\
    \#28 & Co-mining~\cite{sparse_comining} & & ResNet-50 & 1$\times$ & 17.9 & 12.4 & 3.1 & 2.0 & 11.2 & 17.5 & 19.7 \\
    \#29 & Calibrated Teacher~\cite{sparse_calibratedteacher} & & ResNet-50 & 1$\times$ & 16.5 & 11.4 & 3.0 & 3.0 & 10.2 & 16.5 & 17.7 \\
    \midrule
    \#30 & Faster R-CNN & \multirow{3}{*}{5-shot} & ResNet-50 & 1$\times$ & 21.3 & 14.2 & 3.4 & 2.0 & 12.4 & 20.2 & 26.3 \\
    \#31 & Co-mining~\cite{sparse_comining} & & ResNet-50 & 1$\times$ & 23.2 & 15.8 & 3.7 & 3.0 & 14.6 & 22.0 & 28.6 \\
    \#32 & Calibrated Teacher~\cite{sparse_calibratedteacher} & & ResNet-50 & 1$\times$ & 22.1 & 15.2 & 3.7 & 2.9 & 13.2 & 21.3 & 29.2 \\
    \midrule
    \midrule
    \multicolumn{12}{l}{\textit{weakly-supervised:}} \\
    \midrule
    \#33 & P2BNet~\cite{p2bnet} & \multirow{5}{*}{Center Point} & ResNet-50 & 2$\times$ & 4.6 & 0.2 & 0.0 & 1.0 & 1.0 & 3.1 & 0.5 \\
    \#34 & PointOBB-v2~\cite{pointobbv2} & & ResNet-50 & 1.5$\times$ & 16.9 & 7.3 & 0.5 & 1.0 & 8.9 & 13.0 & 18.1 \\
    \#35 & PointOBB-v3~\cite{pointobbv3} & & ResNet-50 & 3$\times$ & 3.0 & 0.1 & 0.0 & 0.0 & 1.0 & 1.9 & 0.2 \\
    \#36 & Point Teacher~\cite{point_teacher} & & ResNet-50 & 1$\times$ & 47.8 & 9.2 & 0.1 & 6.8 & 10.1 & 12.4 & 10.5 \\
    \#37 & Point2Rbox-v3~\cite{point2rboxv3} & & ResNet-50 & 1$\times$ & 46.4 & 20.0 & 2.0 & 1.8 & 15.4 & 24.9 & 42.8 \\
    \midrule
    \midrule
    \multicolumn{12}{l}{\textit{zero-shot:}} \\
    \midrule
    \#38 & Rex-Omni~\cite{rexomni} & \multirow{2}{*}{0-shot} & ViT & 0$\times$ & 4.6 & 2.7 & 0.6 & 0.2 & 1.6 & 5.4 & 7.0 \\
    \#39 & SAM3~\cite{sam3} & & ViT & 0$\times$ & 11.5 & 10.1 & 3.2 & 3.8 & 12.2 & 16.7 & 5.5 \\
	\bottomrule
	\end{tabular}
 }
\end{table*}
\setlength{\tabcolsep}{1.4pt}

In this work, we systematically evaluate a broad range of detection paradigms for small objects.
Through evaluation across fully supervised, label-noise robust, semi-supervised, weakly supervised, sparse-shot, and zero-shot settings, our benchmark provides a comprehensive view of the limitations of different detection paradigms on the SOD task. 
The Implementation Details and Evaluation Metric are discussed in the Appendix.

\begin{figure*}[t]
\centering
\includegraphics[width=\linewidth]{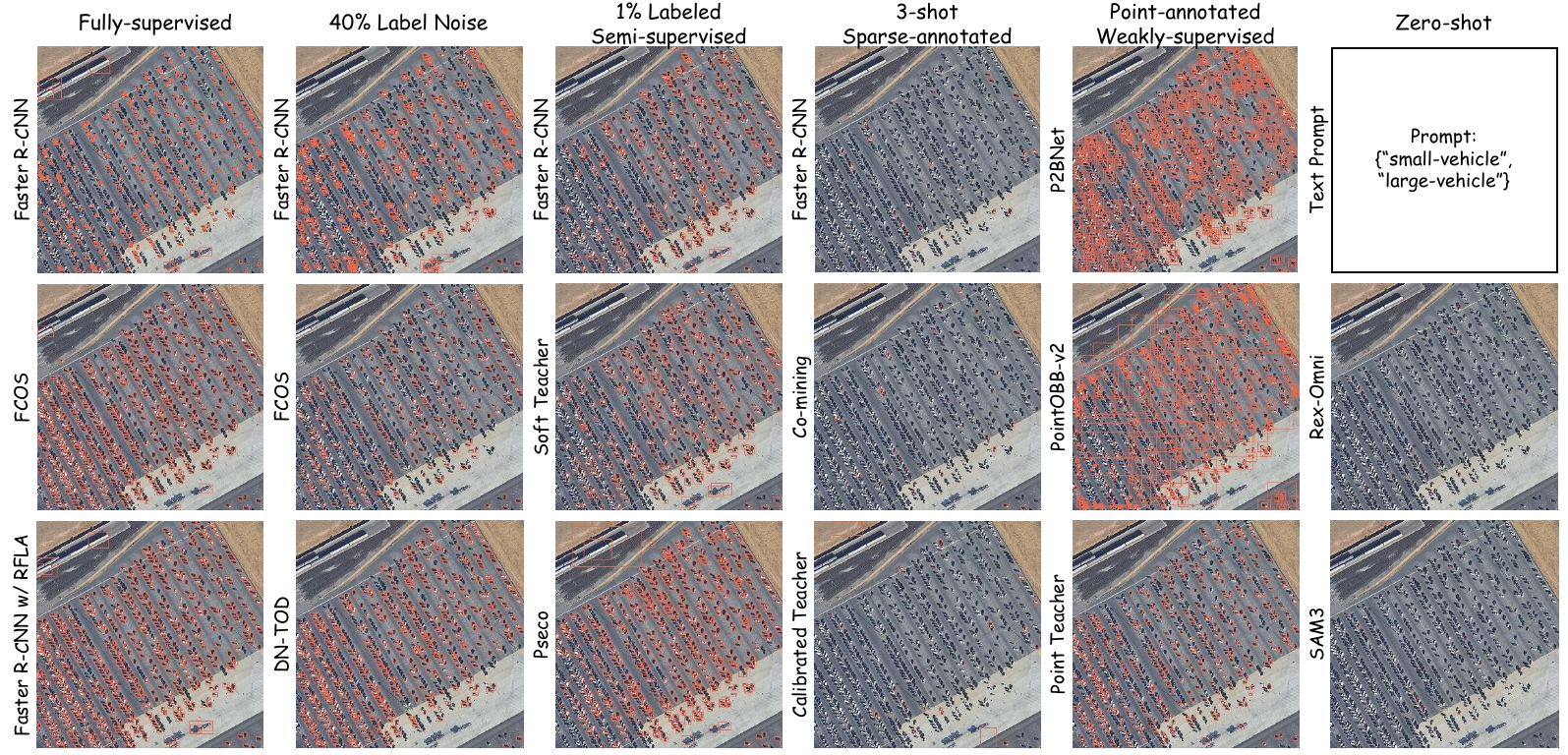}
\caption{Detection results of different detection paradigms on TinySet-9M dataset. Orange boxes denote the detection results with confidence higher than 0.2.}
\label{fig:tinyset_9m_benchmark_detection}
\end{figure*}

\begin{figure*}[t]
\centering
\includegraphics[width=\linewidth]{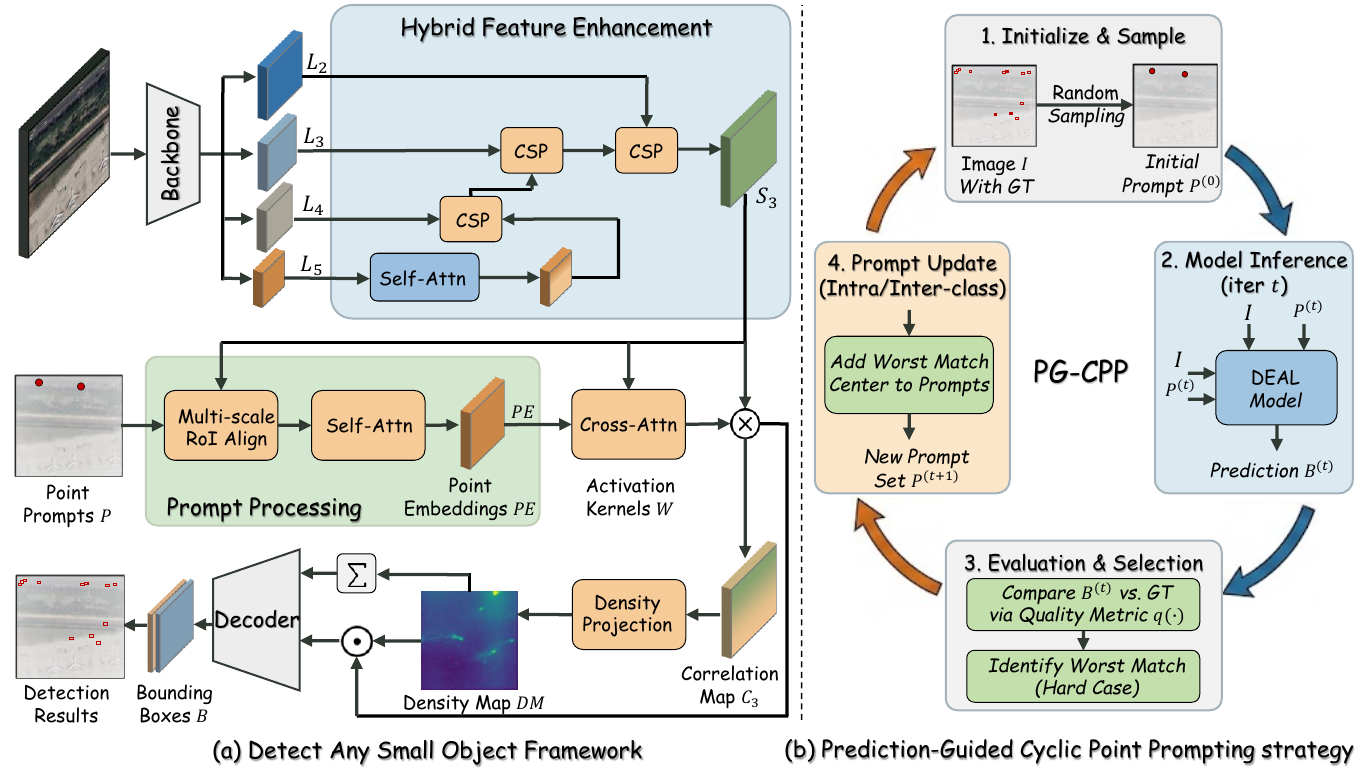}
\caption{Overview of the proposed method. (a) Illustration of the overall architecture of DEAL, which is built upon the RT-DETR framework and extended to support point-prompted small object detection. (b) Illustration of the training pipeline of the proposed Prediction-Guided Cyclic Point Prompting (PG-CPP) strategy.}
\label{fig:method_overview}
\end{figure*}

The experimental results are summarized in Table~\ref{table:bench_tinyset_10m} and Fig.~\ref{fig:tinyset_9m_benchmark_detection}, which present quantitative results and qualitative visualizations of representative methods across different settings. 
We present several noteworthy observations as follows.

\textbf{Performance degradation under diminishing supervision.}
We organize the detection settings according to their supervisory strength, ranging from fully–supervised (\#1-\#6), label–noise–robust (\#7-\#14), and semi-supervised learning (\#15-\#23) to sparse-shot (\#24-\#32), weakly supervised (\#33-\#37), and zero-shot (\#38-\#39) paradigms. As supervision weakens, detection performance consistently deteriorates, dropping from 78.2\% (\#8) and 59.5\% (\#15) to 32.9\% (\#30), and further to 21.3\% (\#36) relative to fully supervised (\#2) performance. This monotonic decline indicates that reduced annotation does not lead to graceful degradation in the SOD, but instead exposes a sharp collapse in localization and recognition capability compared to generic objects. Specifically, under 40\% label noise, small objects achieve only 26.9\% (\#12) of fully supervised performance, whereas generic objects (COCO) retain 49.7\%~\cite{labelnoise_oamil}.
Under the 10\% labeled semi-supervised setting, small objects achieve 69.2\% (\#23) of fully supervised performance, while generic objects reach 78.1\%~\cite{pseco}.

\textbf{Paradigm-specific sensitivity to small-object properties.}
Across different learning paradigms, only methods with task-specific designs for small objects—such as DN-TOD (\#9) and Point Teacher (\#36)—exhibit measurable robustness under reduced supervision. In contrast, most generic label-efficient methods (\#10 and \#33) show limited adaptation, and in sparse-shot settings, some even exhibit negative transfer (\#26). This suggests that label-efficient methods require more carefully designed modules when applied to small object detection.

\textbf{Failure of text-based semantic prompting for small objects.}
Although vision–language models demonstrate strong zero-shot performance on datasets dominated by medium and large objects (For the COCO dataset, the AP50 can reach 86.4\% of fully-supervised methods~\cite{groundingdino}), their effectiveness degrades severely in the small-object regime, achieving only 26.6\% of fully supervised performance (\#39). This failure reflects a misalignment between textual semantics and the visual evidence available for small objects, where blurred boundaries and weak category cues undermine reliable grounding. These results indicate that text prompts alone are insufficient to activate stable category-level perception for small objects.

\textbf{Localization instability under weak supervision.}
Point-level supervision substantially reduces annotation cost but provides limited spatial constraints for precise localization. As a result, localization-sensitive metrics such as $AP_{0.5}$ and $AP_{0.75}$ drop sharply across weakly supervised settings (\#33-\#37). Even methods employing carefully designed two-stage learning strategies, such as Point Teacher (\#36), struggle to recover accurate bounding-box regression, highlighting the amplified sensitivity of small objects to localization noise.

\textbf{Relative robustness of semi-supervised learning.}
Among all evaluated paradigms, semi-supervised learning exhibits comparatively slower performance degradation. With only 1\% labeled data (\#15), semi-supervised methods retain 59.5\% of fully supervised performance. This suggests that leveraging abundant unlabeled data can partially compensate for annotation sparsity, although effective exploitation requires small-object–aware mining and pseudo-label refinement strategies.
\section{Method}
\label{sec:method}

The profound performance degradation revealed in our benchmark underscores a fundamental limitation: the vanishing semantic evidence of small objects constitutes a structural barrier for current detection paradigms. Beyond the scale-induced feature decay, existing detectors are often tethered to predefined categories, severely constraining their generalizability to open-world scenarios and novel object classes. To transcend these limitations, we move away from traditional training-time feature enhancement—which essentially attempts to recover lost signals from noise—and instead propose a paradigm shift toward inference-time semantic activation. By introducing sparse point-based interactions during inference, we inject deterministic category-level cues that act as semantic anchors for elusive targets.
Building on this insight, we formalize \emph{Point-Prompt Small Object Detection (P$^2$SOD)}. This paradigm preserves the rigor of full supervision during training while leveraging sparse point prompts at inference to activate localized category perception, thereby bridging the gap between vanishing visual cues and robust instance localization.
Formally, let the input image be denoted as $I \in \mathbb{R}^{3 \times H \times W}$, and the point prompts be $P \in \mathbb{R}^{k \times 2}, k\in \{1, 2,.., N\}$, where $N$ represents the total number of objects in the image. 
The detection network takes $(I, P)$ as input and outputs a set of bounding boxes $B \in \mathbb{R}^{M \times 4}$ where $M$ denotes the number of objects whose categories match those specified by the point prompts.
Overall, the P$^2$SOD task can be formulated as

\begin{equation}
    B_{c_{i}} = \mathrm{DNN}_{P^2SOD}(I, P_{c_{i}}),
    \label{P2SOD}
\end{equation}
where $c_{i}$ denotes class $i$.

Under the P$^2$SOD formulation, we further develop \emph{DEAL (DEtect Any smalL Object)}, a scalable point-prompted detection framework that integrates prompt-conditioned perception with large-scale representation learning, without relying on heavy multimodal models.

In the following, we will introduce the DEAL architecture and its training paradigm.

\subsection{Detect Any Small Object model}
To enable the interactive P$^2$SOD task, we propose DEAL, a density-aware and prompt-oriented detection framework. DEAL is designed to (i) preserve fine-grained small-object representations, (ii) efficiently integrate point prompts without redundant inference, and (iii) dynamically adapt the decoder behavior to varying prompt conditions.

\textbf{Hybrid Feature Enhancement for Small Object Representation.}
Small objects are highly sensitive to spatial resolution and local details, yet directly incorporating low-level feature maps such as $L_2$ substantially increases computational and memory costs. To balance representation quality and efficiency, we introduce a Hybrid Feature Enhancement (HFE) module inspired by RT-DETR, which enhances small-object features without explicitly introducing high-resolution pyramids. Given backbone features $\{L_2, L_3, L_4, L_5 \}$, we first enhance the highest-level feature $L_5 \in \mathbb{R}^{C \times H/32 \times W/32}$ using a Transformer Encoder to model global dependencies:

\begin{equation}
    \tilde{L_5} = \text{self-attn}(L_5),
    \label{L5}
\end{equation}
where self-attention aggregates long-range contextual information. The globally enhanced feature $\tilde{L_5}$ is then propagated to lower levels through a top-down pathway, while finer spatial details from $L_2$ to $L_4$ are reinforced via a bottom-up fusion process. Both fusion directions are implemented using CSP-based layers:

\begin{equation}
    S_3 = \text{CSP}(\text{Fuse}_{\downarrow}(\tilde{L_5}, L_4, L_3), \text{Fuse}_{\uparrow}(L_2)),
    \label{S3}
\end{equation}
where $S_3 \in \mathbb{R}^{C \times H/8 \times W/8}$ denotes the final enhanced feature map. This hybrid design allows $S_3$ to retain fine-grained spatial details while being guided by global semantic context, making it particularly suitable for small-object detection under limited computational budgets.

\textbf{Point-Guided Density Activation for Prompt–Feature Interaction.}
We decouple visual feature extraction from prompt interaction, performing prompt conditioning exclusively on the enhanced feature map $S_3$. This design is motivated by two critical factors. First, it avoids the redundant backbone computations that would occur if prompts were injected directly, thereby preserving interactive efficiency. More importantly, it acts as a feature activation contrast enhancer. It explicitly amplifies the activation disparity between small objects and the background clutter, compensating for the weak spatial responses that vanish during the progressive downsampling process.
Given a set of point prompts $P={\{p_i\}}_{i=1}^{k}$, where $p_i \in \mathbb{R}^{2}$, we first embed and refine them using a self-attention Transformer to model inter-point relations:

\begin{equation}
    PE=\text{self-attn}(\text{RoI-Align}(P)),
    \label{P'}
\end{equation}
The scale of RoI-Align is $\{8, 16, 32\}$. To activate prompt-relevant regions in the visual feature space, we perform cross-attention between the refined point embeddings $PE$ and the feature map $S_3$:

\begin{equation}
    W=\text{cross-attn}(Q=PE, K=S_3, V=S_3),
    \label{W}
\end{equation}
where $W$ represents prompt-conditioned activation kernels. These kernels are then applied to $S_3$ via convolution to produce a correlation map:
\begin{equation}
    C_3 = S_3 \otimes W,
    \label{C3}
\end{equation}
where $\otimes$ denotes convolution. The resulting feature map $C_3$ highlights regions that are semantically aligned with the point prompts, enabling efficient and flexible prompt-guided feature activation without re-running the backbone.

\textbf{Density-Guided Query Assignment and Decoder Inference.}
In P$^2$SOD, the number of objects corresponding to a given point prompt is unknown and varies across images, making a fixed number of decoder queries suboptimal. To address this, we introduce a density-guided query allocation strategy that dynamically estimates the number of prompt-relevant objects. Specifically, the correlation feature map $C_3$ is projected into a single-channel density map:
\begin{equation}
    DM = \text{Conv}(C_3), DM \in \mathbb{R}^{1 \times H/8 \times W/8},
    \label{density}
\end{equation}
where $DM(x, y)$ represents the likelihood of an object belonging to the prompt category at spatial location $(x, y)$. 
This direct projection transforms prompt-related features into a point density map, which is well-suited to the sparse, point-like representation of small objects and more robustly reflects their true spatial distribution and quantity.
During training, $DM$ is supervised using ground-truth bounding boxes of the corresponding category. The total object count is approximated by spatial integration:
\begin{equation}
    N_{query} = \sum_{x,y}DM(x, y),
    \label{density}
\end{equation}
which is used to determine the number of decoder queries. The prompt-aware feature map is then obtained by modulating $S_3$ with the density map:
\begin{equation}
    \hat{S_3} = S_3 \odot DM,
    \label{S3hat}
\end{equation}
where $\odot$ denotes element-wise multiplication. The modulated feature $\hat{S_3}$, together with the dynamically allocated queries, is fed into the standard RT-DETR decoder without architectural modification, producing the final classification and bounding-box predictions.

\begin{algorithm}[t]
\caption{PG-CPP strategy}
\label{alg:rgcpp}
\small
\KwIn{
Training image $I$; ground-truth objects $\mathcal{G}=\{(b_i,c_i)\}_{i=1}^{N}$; 
detector $\mathrm{DNN}_{P^2SOD}$; number of cycles $T$
}
\KwOut{Optimized detector parameters}

\For{$t = 1$ \KwTo $T$}{
    \tcp{Intra-class cyclic prompting}
    \ForEach{category $c \in \mathcal{C}$}{
        Sample initial point $p_c^{(0)}$ from a GT box of class $c$\;
        \For{$k = 0$ \KwTo $K-1$}{
            $\mathcal{B}_c^{(k)} \leftarrow \mathrm{DNN}_{P^2SOD}(I, p_c^{(k)})$\;
            $b^{*} \leftarrow \arg\min\limits_{b \in \mathcal{B}_c^{(k)}} q(b)$\;
            $p_c^{(k+1)} \leftarrow p_c^{(k)} \cup \mathrm{Center}(b^{*})$\;
        }
    }

    \tcp{Inter-class cyclic prompting}
    Construct prompt set $P^{(0)} = \{p_c^{(0)}\}_{c \in \mathcal{C}}$\;
    \For{$k = 0$ \KwTo $K-1$}{
        $\mathcal{B}^{(k)} \leftarrow \mathrm{DNN}_{P^2SOD}(I, P^{(k)})$\;
        $b^{*} \leftarrow \arg\min\limits_{b \in \mathcal{B}^{(k)}} q(b)$\;
        $P^{(k+1)} \leftarrow P^{(k)} \cup \{\mathrm{Center}(b^{*})\}$\;
    }

    \tcp{Loss computation and optimization}
    Compute $\mathcal{L}_{\text{cls}}, \mathcal{L}_{\text{reg}}, \mathcal{L}_{\text{density}}$\;
    $\mathcal{L} \leftarrow \mathcal{L}_{\text{cls}} + \mathcal{L}_{\text{reg}} + \lambda \mathcal{L}_{\text{density}}$\;
    Backpropagate $\mathcal{L}$ and update model parameters\;
}
\end{algorithm}

\textbf{Training Objective.}
The overall training objective of DEAL jointly optimizes prompt-aware density estimation and object detection. In addition to the standard classification loss $L_{cls}$ and bounding-box regression loss $L_{reg}$, we introduce a density supervision loss $L_{density}$ to guide the learning of $DM$. The density loss is implemented using a focal loss formulation to handle the severe imbalance between object and background regions.
\begin{equation}
    L_{density} = \text{FL}(DM, DM^{gt}),
    \label{density_loss}
\end{equation}
where $DM^{gt}$ is constructed from ground-truth boxes of the prompt category. FL($\cdot$) denotes the focal loss~\cite{Focal-Loss_2017_ICCV}.
The final loss is defined as:
\begin{equation}
    L = L_{cls}+L_{reg}+\lambda L_{density},
    \label{losses}
\end{equation}
with $\lambda$ controlling the contribution of the density supervision. This joint optimization enables DEAL to learn robust prompt-conditioned detection while preserving the advantages of fully supervised training.

\begin{figure*}[t]
\centering
\includegraphics[width=\linewidth]{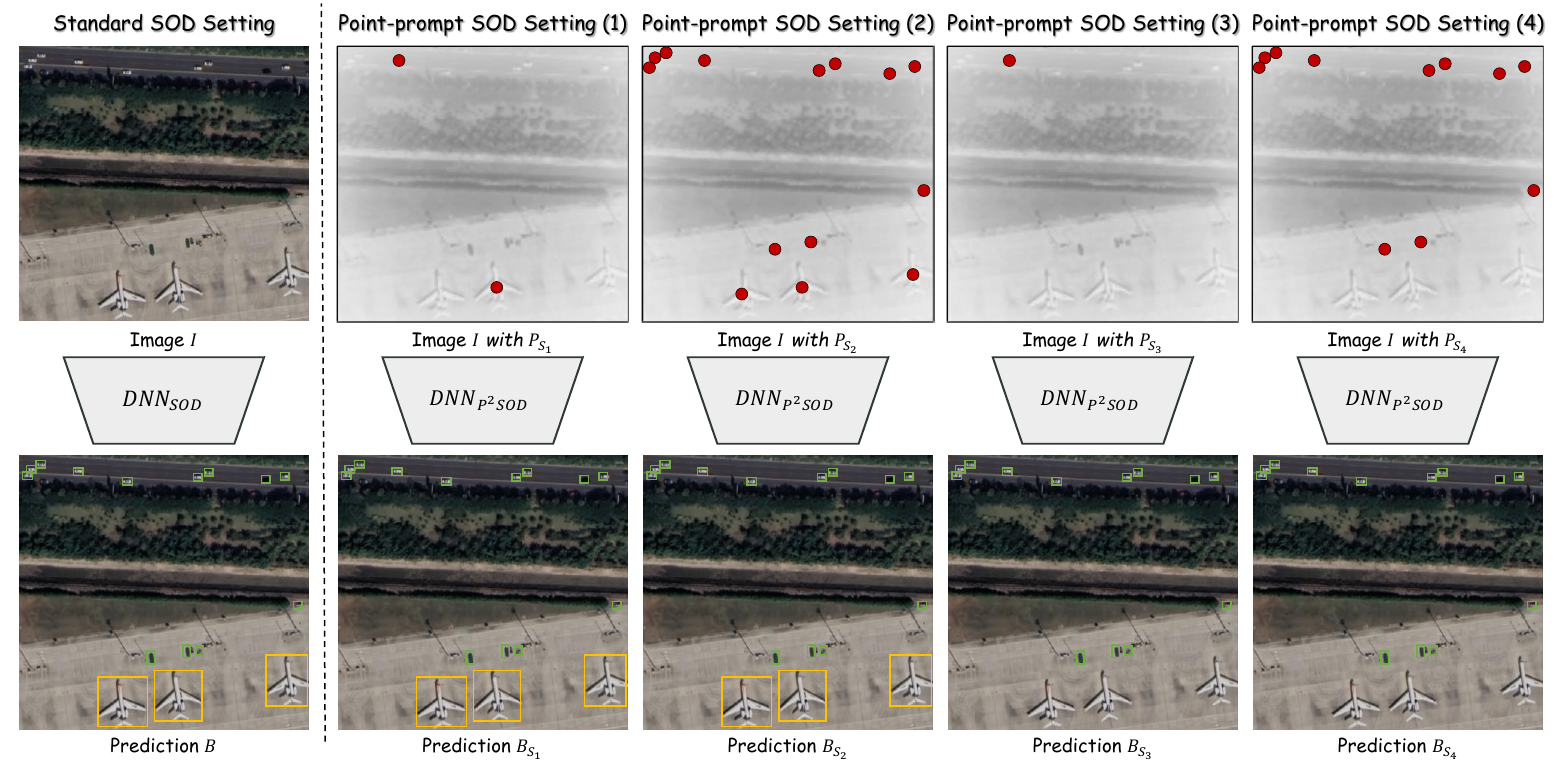}
\caption{Inference settings for the P$^2$SOD task. Setting (1) uses one point prompt per category in each image; Setting (2) assigns one point prompt to each object instance; Setting (3) randomly selects a single category and uses one point from that category as the prompt; Setting (4) randomly selects a category and uses point prompts from all its instances.}
\label{fig:experiment_settings}
\end{figure*}

\subsection{Prediction-guided Cyclic Point Prompting Strategy}
Training a P$^2$SOD detector requires the model to generalize across diverse point-prompt configurations; however, naively sampling random points during training fails to expose the detector to hard cases and often leads to prompt overfitting. To address this issue, we propose a Prediction-guided Cyclic Point Prompting (PG-CPP) strategy, which adaptively refines point prompts based on the model’s current detection performance. Given a training image $I$ with ground-truth objects $\mathcal{G}={\{ (b_i, c_i) \}}_{i=1}^{N}$, where $b_i$ and $c_i$ denote the bounding box and category label, we construct two complementary cyclic prompting schemes: an intra-class cycle and an inter-class cycle. In the intra-class cycle, for each category $c$, an initial point prompt $p_c^{(0)}$ is randomly sampled from the interior of a ground-truth box belonging to that category. The detector then produces a set of predictions $B_c^{(t)} = DNN_{P^2SOD}(I, p_c^{(t)})$ at iteration $t$. Among all matched predictions, we identify the worst-performing object according to a matching quality metric $q(\cdot)$, defined by classification confidence and localization accuracy, and select its center and reintroduce it as an additional prompt:
\begin{equation}
    p_c^{(t+1)}=\arg \min_{b\in B_c^{(t)}}q(b) \cup p_c^{(t)}.
    \label{pt+1}
\end{equation}

This process forces the detector to iteratively focus on hard-to-detect instances within the same category. In the inter-class cycle, we sample one point from each category to form a prompt set $P^{(t)}={\{p_c^{(t)}\}}_{c\in C}$, where $C$ denotes the set of categories present in the image. The detector predicts a joint result $B^{(t)}=DNN_{P^2SOD}(I, P^(t))$, from which the worst-matched object across all categories is selected and reintroduced as an additional prompt:
\begin{equation}
    P^{(t+1)}=\arg \min_{b\in B^{(t)}}q(b) \cup P^{(t)}.
    \label{Pt+1}
\end{equation}

By alternating between intra-class and inter-class cycles, PG-CPP systematically exposes the model to challenging prompt–object associations and enforces robustness to varying prompt configurations. This prediction-guided prompting mechanism enables the detector to learn adaptive prompt-conditioned behaviors and significantly improves generalization at inference time, where only sparse point prompts are available.
\section{Experiments}
\label{sec:experiments}
\subsection{Datasets and Implementations Details}
\label{sec:experiments_implementation}

\setlength{\tabcolsep}{3pt}
\begin{table*}[t]
\centering
\caption{Main results of fully-supervised, point-supervised, zero-shot (box-prompt), zero-shot (text-prompt) methods, and our proposed DEAL on TinySet-9M (class-agnostic). For the training schedule, 1$\times$ denotes 3 epochs. All experiments are run on a computer with an NVIDIA RTX 4090 (24 GB) GPU. We use FP32 with 1024 $\times$ 1024 inputs. Our proposed DEAL is conducted under setting (1).}
\label{table:bench_daso}
\renewcommand{\arraystretch}{0.7}
\renewcommand{\tabcolsep}{1.0mm}
\resizebox{0.95\linewidth}{!}{
\begin{tabular}{l|c|c|ccc|cccc}
	\toprule
	Method & Backbone & Schedule & $\rm{AP_{0.25}}$ & $ \rm{AP_{0.5}}$ & $\rm{AP_{0.75}}$ & $\rm{AP_{vt}}$ & $\rm{AP_{t}}$ & $\rm{AP_{s}}$ & $\rm{AP_{m}}$ \\
	\midrule
    \multicolumn{10}{l}{\textit{fully-supervised:}} \\
    \midrule
    RetinaNet~\cite{Focal-Loss_2017_ICCV} & ResNet-50 & 1$\times$ & 56.7 & 37.8 & 10.6 & 6.5 & 25.0 & 49.7 & 83.3 \\
    Faster R-CNN~\cite{Faster-R-CNN_2015_NIPS} & ResNet-50 & 1$\times$ & 51.9 & 43.2 & 17.8 & 5.0 & 30.5 & 59.0 & 84.5 \\
    FCOS~\cite{FCOS_2022_TPAMI} & ResNet-50 & 1$\times$ & 66.3 & 49.5 & 17.1 & 13.8 & 41.1 & 62.3 & 83.7 \\
    RT-DETR~\cite{rtdetr} & ResNet-50 & 1$\times$ & 57.2 & 43.1 & 12.1 & 11.3 & 35.3 & 54.7 & 74.1 \\
    HS-FPN (Faster R-CNN)~\cite{tod_hsfpn} \qquad \qquad \qquad & ResNet-50 & 1$\times$ & 51.8 & 43.0 & 17.5 & 5.9 & 31.2 & 59.1 & 83.0 \\ 
    RFLA (Faster R-CNN)~\cite{rfla_2022_eccv} & ResNet-50 & 1$\times$ & 66.9 & 54.6 & 20.8 & 15.3 & 46.6 & 67.5 & 88.4 \\ 
    \midrule
    \midrule
    \multicolumn{10}{l}{\textit{point-supervised:}} \\
    \midrule
    P2BNet~\cite{p2bnet} & ResNet-50 & 1$\times$ & 4.6 & 0.2 & 0.0 & 1.0 & 1.0 & 3.1 & 0.5 \\ 
    PointOBB-v2~\cite{pointobbv2} & ResNet-50 & 1$\times$ & 16.9 & 7.3 & 0.5 & 1.0 & 8.9 & 13.0 & 18.1 \\ 
    PointOBB-v3~\cite{pointobbv3} & ResNet-50 & 1$\times$ & 3.0 & 0.1 & 0.0 & 0.0 & 1.0 & 1.9 & 0.2 \\ 
    Point Teacher~\cite{point_teacher} & ResNet-50 & 1$\times$ & 47.8 & 9.2 & 0.1 & 6.8 & 10.1 & 12.4 & 10.5 \\ 
    Point2Rbox-v3~\cite{point2rboxv3} & ResNet-50 & 1$\times$ & 46.4 & 20.0 & 2.0 & 1.8 & 15.4 & 24.9 & 42.8 \\ 
    \midrule
    \midrule
    \multicolumn{10}{l}{\textit{zero-shot (one-box-prompt):}} \\
    \midrule
    Rex-Omni~\cite{rexomni} & ViT & 0$\times$ & 3.8 & 2.1 & 0.4 & 0.1 & 1.7 & 4.4 & 3.7 \\
    SAM3~\cite{sam3} & ViT & 0$\times$ & 20.5 & 18.6 & 9.6 & 6.6 & 16.1 & 27.6 & 21.1 \\
    \midrule
    \midrule
    \multicolumn{10}{l}{\textit{zero-shot (text-prompt):}} \\
    \midrule
    Rex-Omni~\cite{rexomni} & ViT & 0$\times$ & 4.6 & 2.7 & 0.6 & 0.2 & 1.6 & 5.4 & 7.0 \\
    SAM3~\cite{sam3} & ViT & 0$\times$ & 11.5 & 10.1 & 3.2 & 3.8 & 12.2 & 16.7 & 5.5 \\
    \midrule
    \midrule
    \multicolumn{10}{l}{\textit{Our Proposed method:}} \\
    \midrule
    \rowcolor{gray!20} DEAL & Swin-T & 1x & 71.4 & 53.8 & 15.9 & 6.9 & 43.3 & 71.6 & 69.2 \\
    \rowcolor{gray!20} DEAL* & Swin-T & 2x & 71.9 & 55.2 & 18.2 & 7.9 & 45.2 & 73.0 & 71.1 \\
	\bottomrule
	\end{tabular}
}
\end{table*}
\setlength{\tabcolsep}{1.4pt}

\textbf{Datasets.} 
All main experiments and ablation studies are conducted on TinySet-9M, which is divided into training, validation, and test splits. Models are trained on the training set, validated on the validation set, and evaluated on the test set. 
To evaluate generalization to unseen categories and unseen datasets, we conduct experiments on two representative remote sensing benchmarks, DIOR~\cite{DIOR_2019_ISPRS} and DOTA-v2.0~\cite{DOTA2.0_2021_pami}. In addition, to assess whether the backbone pre-trained on TinySet-9M can better capture small-object characteristics and transfer this capability to other datasets, we further perform fine-tuning experiments on COCO~\cite{COCO_2014_ECCV}, TinyPerson~\cite{TinyPerson_2020_WACV}, RGBTDronePerson~\cite{RGBTDronePerson}, and xView~\cite{xview_2018_arXiv}.

Since the P$^2$SOD task requires both an image and point prompts as inputs at inference time, we \textbf{design multiple evaluation settings} to systematically examine detection performance under different point-prompt configurations, as shown in Fig.~\ref{fig:experiment_settings}.
\textbf{Setting (1)} constructs a point set $P_{S_1}$ by randomly selecting one instance per category in each image and using the center of its bounding box as the point prompt; this setting primarily evaluates recall and localization accuracy.
\textbf{Setting (2)}  forms a point set $P_{S_2}$ by using the center point of every object instance in the image as a prompt, which is used to assess localization accuracy and the responsiveness of the predicted density maps.
\textbf{Setting (3)} constructs a point set $P_{S_3}$ by randomly selecting one category and one instance from that category per image, and using its center as the point prompt; this setting evaluates recall and category discrimination under sparse prompting.
\textbf{Setting (4)} builds a point set $P_{S_4}$ by randomly selecting one category and using the center points of all its instances as prompts, which is designed to evaluate localization accuracy, density-map responsiveness, and category-specific discrimination.

\textbf{Implementation details.} 
All experiments are conducted under the same settings in Section~\ref{sec:benchmark} to ensure fair comparison and reproducibility. Models are trained on the training split and evaluated on the validation split for DOTA-v2.0, while inference on TinyPerson, RGBTDronePerson, and xView is performed at a resolution of 1024 $\times$ 1024. Experiments are carried out on a single NVIDIA RTX 4090 GPU using the MMDetection framework, with all unspecified hyperparameters set to their default values.

\setlength{\tabcolsep}{3pt}
\begin{table*}[t]
\centering
\caption{ The cross-dataset evaluation results on DIOR, DOTA-v2.0 test set. The results of DOTA-v2.0 are all based on horizontal detection boxes. GeoImageNet, Sentinel-2, TOV-NI, TOV-R, FMoW, SatlasPretrain, MillionAID, RingMoPretrain, and multi-modal RSI are remote sensing datasets. Our proposed DEAL is conducted under setting (1). ($\cdot$) denotes $\rm{AP_{small}}$. } 
\label{table:bench_daso_finetune}
\renewcommand{\arraystretch}{0.85}
\renewcommand{\tabcolsep}{1.0mm}
\resizebox{0.95\linewidth}{!}{
\begin{tabular}{l|c|c|cc|cc}
	\toprule
	\multirow{2}{*}{Method} & \multirow{2}{*}{Backbone} & \multirow{2}{*}{Pre-Training Data} & \multicolumn{2}{c|}{w/o fine-tuning} & \multicolumn{2}{c}{w/ fine-tuning} \\
      &  &  & DIOR & DOTA-v2.0 & DIOR & DOTA-v2.0 \\
	\midrule
    \multicolumn{7}{l}{\textit{Generic Object Detection}} \\
    \midrule
    RetinaNet~\cite{Focal-Loss_2017_ICCV} & ResNet-50 & - & - & - & 66.1 & 51.6 \\
    Faster R-CNN~\cite{Faster-R-CNN_2015_NIPS} & ResNet-50 & - & - & - & 52.1 & 54.8 \\
    FCOS~\cite{FCOS_2019_ICCV} & ResNet-50 & - & - & - & 71.9 & 54.3 \\
    \midrule
    GASSL~\cite{GASSl} & ResNet-50 & - & - & - & 67.4 & - \\
    CACO~\cite{CACO} & ResNet-50 & Sentinel-2 & - & - & 66.9 & - \\
    TOV~\cite{TOV} & ResNet-50 & TOV-NI,TOV-R & - & - & 70.2 & - \\
    Scale-MAE~\cite{Scale-MAE} & ViT-L & FMoW & - & - & 73.8 & - \\
    SatLas~\cite{SatLas} & Swin-B & SatlasPretrain & - & - & 74.1 & - \\
    RingMo~\cite{RingMo} & Swin-B & RingMoPretrain & - & - & 75.9 & - \\
    SkySense~\cite{SkySense} & Swin-H & multi-modal RSI & - & - & 78.7 & - \\
    MTP~\cite{MTP} & Swin-H & MillionAID & - & - & 81.1 & - \\
    \midrule
    \multicolumn{7}{l}{\textit{Visual-Language Models}} \\
    \midrule
    Rex-Omni~\cite{rexomni} (Text Prompt) & ViT & - & 25.5 (17.6) & 10.6 (10.6) & - & - \\
    Rex-Omni~\cite{rexomni} (Visual Prompt) & ViT & - & 18.8 (15.4) & 7.2 (8.2) & - & - \\
    SAM3~\cite{sam3} (Text Prompt) & ViT & SA-Co & 33.9 (33.9) & 21.2 (17.3) & - & - \\
    SAM3~\cite{sam3} (Visual Prompt) & ViT & SA-Co & 80.0 (77.6) & 43.3 (44.0) & - & - \\
    \midrule
    \multicolumn{7}{l}{\textit{Point-prompt Small Object Detection}} \\
    \midrule
    \rowcolor{gray!20} DEAL & Swin-T & TinySet-9M & 54.3 (81.8) & 55.9 (77.7) & 84.9 & 71.5 \\
	\bottomrule
	\end{tabular}
}
\end{table*}
\setlength{\tabcolsep}{1.4pt}

\subsection{Main Experiments}
The performance on the TinySet-9M test set is reported in Table~\ref{table:bench_daso}, where we compare DEAL with closely related fully supervised detectors, point-supervised methods, and zero-shot approaches, including both visual- and text-prompted models. Under full supervision, RT-DETR achieves an $\rm{AP_{0.5}}$ of 43.1, while the proposed DEAL improves this result by 10.7 points, demonstrating a substantial performance gain. This improvement stems not only from the architectural optimizations tailored for small-object detection but also from the proposed inference paradigm, which leverages additional point prompts to provide informative guidance during detection.
For fair comparison, we further evaluate state-of-the-art zero-shot methods, including Rex-Omni and SAM3. SAM3 achieves an $\rm{AP_{0.5}}$ of 18.6 under visual prompting and 10.1 under text prompting, indicating strong zero-shot capability. However, a large performance gap remains compared to supervised and point-prompted approaches, which can be attributed to the limited coverage of object categories and concepts encountered during SAM3’s pretraining. Similarly, Rex-Omni fails to achieve competitive accuracy on TinySet-9M. Beyond the category coverage issue, Rex-Omni adopts an autoregressive detection paradigm based on a Qwen model~\cite{qwen25vl}, which struggles in scenes containing a large number of densely distributed objects. Moreover, we observe a noticeable performance drop for Rex-Omni under visual prompting, primarily due to a high rate of false positives, which further degrades its detection accuracy.
In contrast, our method, using only a single point prompt per category (Setting (1)), achieves highly competitive performance, which empirically validates the effectiveness of the proposed paradigm for detecting small and densely distributed objects.

\subsection{Generalization Experiments}
To evaluate generalization to unseen categories and unseen datasets, we conduct experiments on two widely used remote-sensing benchmarks, DIOR and DOTA-v2.0. For consistency, horizontal bounding boxes are adopted as the evaluation format for both datasets. The quantitative results are summarized in Table~\ref{table:bench_daso_finetune}, where we compare three representative detection paradigms: generic object detectors, vision–language models (VLMs), and the proposed point-prompt small object detection framework.
Despite having no exposure to the target datasets during training, our method achieves strong detection performance across both benchmarks, obtaining $\rm{AP_{0.5}}$ scores of 54.5 and 59.7 on DIOR and DOTA-v2.0, respectively. The relatively lower performance on DIOR can be attributed to its object scale distribution, as DIOR mainly consists of large objects. Specifically, the average object size in DIOR is 119.3 pixels, whereas nearly all annotations in the pretraining dataset TinySet-9M are smaller than 32 pixels. Accordingly, we further report small-object-specific performance in Table~\ref{table:bench_daso_finetune}, where our method demonstrates consistently strong results. We further prove that fine-tuning on the training splits of DIOR and DOTA-v2.0 would further improve performance under these conditions (from 54.3 to 84.9 on DIOR).
We additionally analyze large-scale vision–language models, including SAM3 and Rex-Omni, whose performance trends are consistent with those observed on TinySet-9M. While both models exhibit strong generalization due to large-scale pretraining and cross-modal representations, their effectiveness remains limited in scenes dominated by small and densely distributed objects. In particular, SAM3 performs reasonably under visual prompting but degrades noticeably under text prompting, indicating that weak visual cues and ambiguous boundaries hinder reliable vision–language alignment for small objects. Rex-Omni, which adopts an autoregressive detection paradigm, similarly struggles in dense scenarios, where error accumulation and excessive false positives significantly impair detection accuracy. These results further suggest that, despite their impressive generalization ability, handling small and dense objects in a fully zero-shot manner remains a fundamental challenge for current VLMs.

\subsection{Ablation Study}
We conduct five ablation studies to systematically validate the effectiveness of the proposed method from complementary perspectives.

\textbf{Effect of inference settings.}
Increasing the number of point prompts consistently improves detection performance. As shown in Table~\ref{tab:ablations}(a), this gain is particularly evident in single-category scenarios, where multiple prompts provide stronger cues for distinguishing target objects from clutter, indicating the importance of sufficient point-level guidance.

\textbf{Effect of the number of point prompts.}
Performance improves with more point prompts but quickly saturates. Table~\ref{tab:ablations}(b) shows that although additional prompts enhance accuracy, the gain becomes marginal beyond a certain number, while computational cost increases. In practice, 3–10 points provide a good trade-off between accuracy and efficiency.

\textbf{Effect of backbone architectures.}
Backbones with stronger global modeling capability yield better performance. As shown in Table~\ref{tab:ablations}(c), PVTv2 and Swin Transformer outperform ResNet-50, while the frozen DINOv3 backbone performs worse due to limited adaptability to prompt-conditioned detection. We adopt Swin Transformer as the default backbone for its favorable balance.

\textbf{Effect of module combinations.}
Each component contributes positively to the overall performance. As shown in Table~\ref{tab:ablations}(d), Point Embedding (PE) is essential for enabling point prompts, Hybrid Feature Enhancement (HFE) improves small-object representation, and DGQA significantly enhances category-level detection. The PG-CPP strategy provides additional gains, confirming the effectiveness of adaptive prompt refinement.

\begin{figure}[t]
\centering
\includegraphics[width=0.9\linewidth]{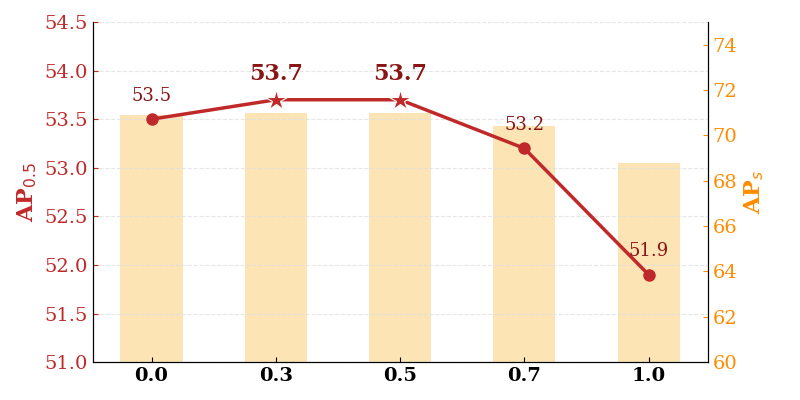}
\caption{Different point prompt position during inference under setting (3).}
\label{fig:ablation_point_position}
\end{figure}

\textbf{Robustness to point location.}
The proposed method is robust to variations in point location. As shown in Fig.~\ref{fig:ablation_point_position}, moderate perturbations of point positions slightly improve performance, while larger perturbations lead to minor degradation, demonstrating stability under imprecise interactions.

\begin{table*}[t]\vspace{-3mm}
\centering
\caption{Ablations. We train on TinySet-9M \texttt{train set}, test on \texttt{test set}. 
HFE, PE, and DGQA represent Hybrid Feature Enhancement, Point Embedding, and Point-Guided Density Activation, respectively.
Gray row means the default setting. }
\subfloat[Different inference setting results. \label{tab:ablation_settings}]{
\tablestyle{8pt}{1.25}\begin{tabular}{c | c c c c c}  
	\toprule
    Settings & AP$_{0.5}$ & AP$_{vt}$ & AP$_{t}$ & AP$_{s}$ & AP$_{m}$ \\
    \midrule
    (1) & 53.8 & 6.9 & 43.4 & 71.6 & 69.2 \\
    (2) & 56.9 & 8.8 & 46.8 & 74.8 & 72.4 \\
    \midrule
    (3) & 53.5 & 5.0 & 42.7 & 70.9 & 66.8 \\
    (4) & 58.2 & 7.2 & 48.2 & 75.1 & 70.8 \\
    \bottomrule
\end{tabular}}\hspace{3mm}
\subfloat[Different number of point prompts under setting (3).
\label{tab:ablation_settings(3)_num}]{
\tablestyle{8pt}{1.35}\begin{tabular}{c | c c c c c}  
	\toprule
    Num. of Points & AP$_{0.5}$ & AP$_{vt}$ & AP$_{t}$ & AP$_{s}$ & AP$_{m}$ \\
    \midrule
    1 & 53.5 & 5.0 & 42.7 & 70.9 & 66.8 \\
    2 & 55.3 & 5.5 & 44.8 & 72.6 & 69.1 \\
    3 & 55.8 & 5.9 & 45.3 & 73.1 & 69.7 \\
    $\infty$ & 58.2 & 7.2 & 48.2 & 75.1 & 70.8 \\
    \bottomrule
\end{tabular}} \\
\subfloat[Different backbone results under setting (3). \label{tab:ablation_backbone}]{
\tablestyle{9pt}{1.25}\begin{tabular}{c | c c c c c}  
	\toprule
    Backbone & AP$_{0.5}$ & AP$_{vt}$ & AP$_{t}$ & AP$_{s}$ & AP$_{m}$ \\
    \midrule
    ResNet50~\cite{backbone_resnet} & 47.7 & 2.6 & 33.4 & 66.8 & 65.7 \\
    PVT-v2~\cite{backbone_pvtv2} & 51.7 & 4.1 & 42.5 & 71.1 & 66.5 \\
    DINO-v3~\cite{backbone_dinov3} & 24.1 & 2.1 & 19.8 & 34.8 & 21.1 \\
    \rowcolor{gray!20} Swin-T~\cite{backbone_swimt} & 53.5 & 5.0 & 42.7 & 70.9 & 66.8 \\
    \bottomrule
\end{tabular}}\hspace{3mm}
\subfloat[Individual effectiveness of components in DEAL.
\label{tab:ablation_components}]{
\tablestyle{10pt}{1.05}\begin{tabular}{c c c | c | c}  
	\toprule
    PE & HFE & DGQA & PG-CPP &  AP$_{0.5}$ \\
    \midrule
     &  &  &  & 0.0 \\
    \checkmark &  &  &  & 4.6 \\
    \checkmark & \checkmark &  &  & 39.2 \\
    \checkmark & \checkmark & \checkmark &  & 52.0 \\
    \rowcolor{gray!20} \checkmark & \checkmark & \checkmark & \checkmark & 53.5 \\
    \bottomrule
\end{tabular}} 
\label{tab:ablations}\vspace{-3mm}
\end{table*}

\setlength{\tabcolsep}{3pt}
\begin{table}[t]
\centering
\caption{Training and inference memory consumption and runtime overhead with batch\_size=1.}
\label{table:daso_training_inference_consumption}
\renewcommand{\arraystretch}{0.8}
\renewcommand{\tabcolsep}{1.0mm}
\resizebox{0.95\linewidth}{!}{
\begin{tabular}{c | c c c c}  
	\toprule
    \multirow{2}{*}{Methods} & Training & Training & Inference & Inference \\
     & time (ms) & memory (MB) & time (ms) & memory (MB) \\
    \midrule
    RT-DETR & 98.8 & 9430.0 & 28.1 & 490.2 \\
    Rex-Omni & -- & -- & 4026.9 & 20264.0 \\
    SAM3 & -- & -- & 169.3 & 4354.8 \\
    \rowcolor{gray!20} DEAL & 320.2 & 11036.0 & 40.8 & 581.1 \\
    \bottomrule
	\end{tabular}
 }
\end{table}
\setlength{\tabcolsep}{1.4pt}

\subsection{Training and Inference Efficiency}
We compare the training efficiency of the proposed method with the baseline detector RT-DETR, with results reported in Table~\ref{table:daso_training_inference_consumption}. Due to the introduction of the cyclic PG-CPP optimization strategy, the training time increases moderately. However, the GPU memory consumption during training remains low and comparable to the baseline.
More importantly, we evaluate inference efficiency by comparing our method with large-scale models, including Rex-Omni and SAM3, in terms of inference latency and memory usage. Our proposed method achieves the fastest inference speed, with a latency of 40.8\,ms per image and a GPU memory footprint of 581.1\,MB. These results demonstrate that the proposed framework offers an efficient and scalable solution for point-prompt small object detection without incurring the high computational overhead typically associated with large vision–language models.


\setlength{\tabcolsep}{3pt}
\begin{table}[t]
\centering
\caption{Performance of the trained feature extractors when transferred to other object detection tasks. In these experiments, FCOS is adopted as the detection framework, and all models are trained for 3 epochs.}
\label{table:daso_zero_shot}
\renewcommand{\arraystretch}{0.8}
\renewcommand{\tabcolsep}{1.0mm}
\resizebox{0.95\linewidth}{!}{
\begin{tabular}{l|c|ccccc}
	\toprule
	Dataset & Pre-training Data & $ \rm{AP_{0.5}}$ & $\rm{AP_{vt}}$ & $\rm{AP_{t}}$ & $\rm{AP_{s}}$ & $\rm{AP_{m}}$ \\
	\midrule
    \multirow{2}{*}{COCO} & ImageNet & 50.8 & 7.6 & 25.5 & 37.1 & 60.0 \\  
     & TinySet-9M & 51.0 & 7.9 & 25.3 & 37.1 & 60.5 \\  
    \midrule
    \multirow{2}{*}{TinyPerson} & ImageNet & 8.9 & 0.5 & 2.0 & 9.7 & 28.2 \\  
     & TinySet-9M & 9.1 & 0.5 & 2.0 & 9.8 & 29.8 \\  
    \midrule
    \multirow{2}{*}{xView} & ImageNet & 30.6 & 2.2 & 9.7 & 23.6 & 32.1 \\  
     & TinySet-9M & 31.0 & 4.0 & 11.8 & 25.0 & 32.7 \\   
    \midrule
    \multirow{2}{*}{RGBT} & ImageNet & 23.5 & 9.8 & 21.4 & 32.4 & 22.3 \\  
     & TinySet-9M & 24.8 & 15.1 & 22.9 & 33.8 & 16.7 \\   
	\bottomrule
	\end{tabular}
 }
\end{table}
\setlength{\tabcolsep}{1.4pt}

\begin{figure*}[t]
\centering
\includegraphics[width=\linewidth]{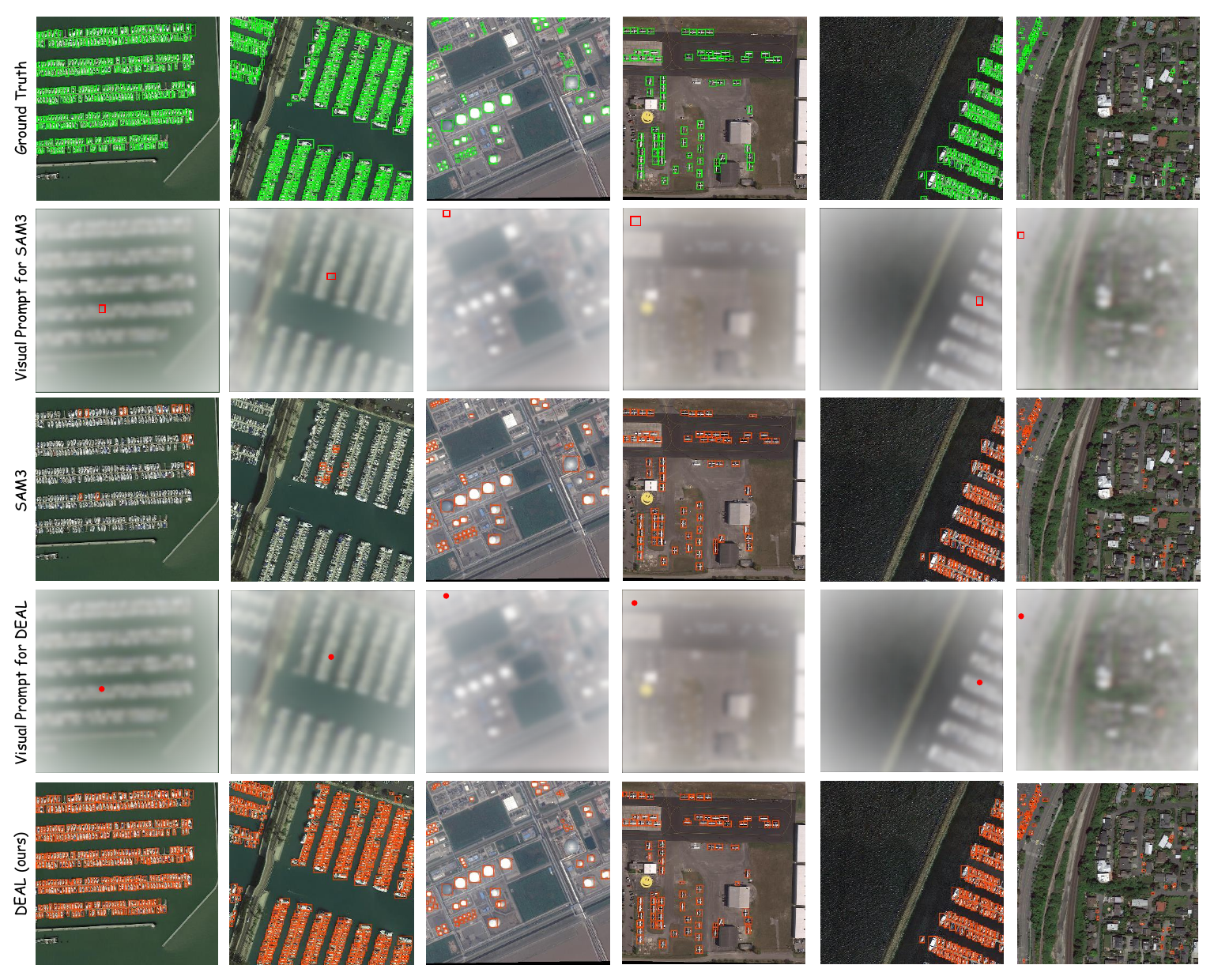}
\caption{Detection results of zero-shot methods SAM3 and our proposed DEAL on DOTA-v2.0 dataset. Green boxes, red boxes, red points, and orange boxes denote the \textit{gt}, box visual prompts, point visual prompts, and detection results, respectively. }
\label{fig:detection_zero_shot_dotav2}
\end{figure*}

\subsection{Results on More Datasets}
\begin{figure*}[t]
\centering
\includegraphics[width=\linewidth]{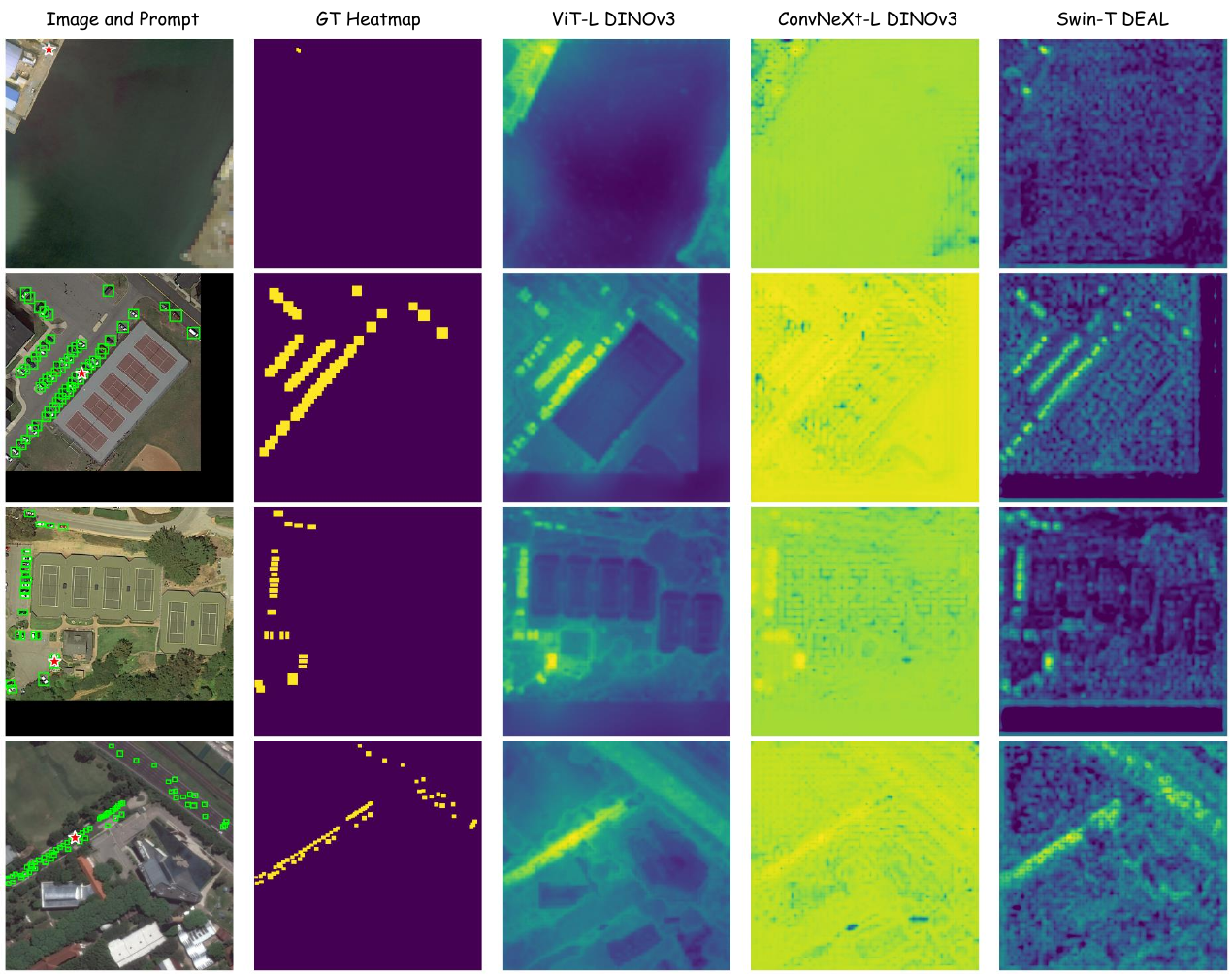}
\caption{Unseen sceneries feature association visualization. In this figure, green bounding boxes denote ground-truth annotations, and red stars indicate the point prompts. Each feature map visualizes the cosine similarity between the feature at every spatial location and the feature at the prompted point location. }
\label{fig:feature_map}
\end{figure*}

Furthermore, we show that backbones pre-trained on TinySet-9M acquire stronger small-object representations and improved object association capability on unseen datasets and categories. To validate this claim, we conduct both quantitative and qualitative evaluations.
For the quantitative analysis, we fine-tune the TinySet-9M–pretrained backbone and an ImageNet-pretrained counterpart on several unseen datasets, including COCO, TinyPerson, xView, and RGBTDronePerson. The results, reported in Table~\ref{table:daso_zero_shot}, consistently demonstrate improved performance on small objects when using the TinySet-9M–pretrained backbone. Notably, the performance gains become more pronounced as the target dataset size decreases, indicating that large-scale small-object pretraining enables faster and more effective adaptation under limited data.
For the qualitative analysis, we visualize the correlation-aware feature maps on DOTA-v2.0, as shown in Fig.~\ref{fig:feature_map}. Compared to the state-of-the-art DINOv3 feature extractor, our backbone exhibits stronger and more coherent feature associations for small objects, highlighting its superior ability to capture fine-grained small-object relationships.

\begin{figure*}[t]
\centering
\includegraphics[width=\linewidth]{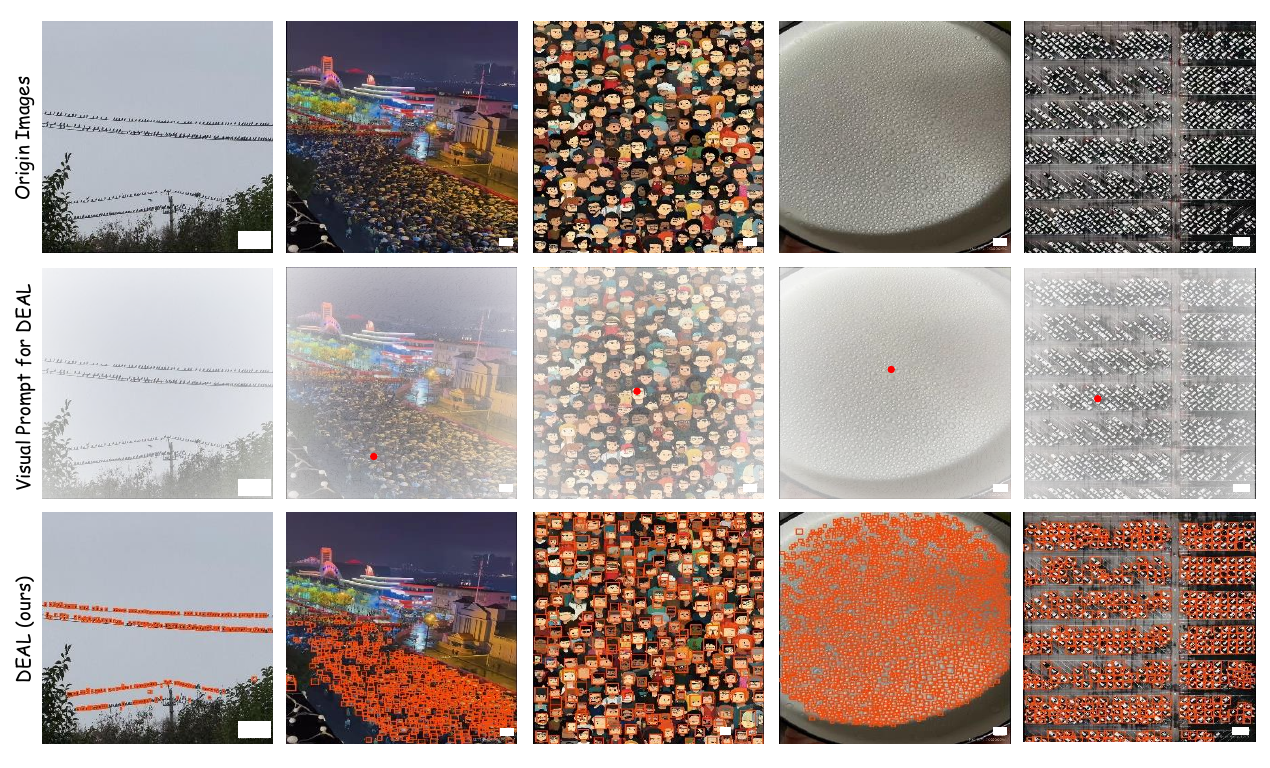}
\caption{Detection results on images collected from the Internet.
All images are resized to a resolution of 1024 $\times$ 1024. None of these images belong to the training datasets, and they exhibit unseen categories or data distributions, such as cartoon images and surveillance camera footage. Red points and orange boxes denote the point visual prompts and the corresponding detection results, respectively. }
\label{fig:detection_zero_shot_redbook}
\end{figure*}

\subsection{Visualization}
We present qualitative visualizations on DOTA-v2.0 to further analyze the behavior of the proposed method. On DOTA-v2.0, we assess generalization under distribution shift. For comparison, we include the state-of-the-art model SAM3, evaluated under visual-prompt–based inference.
As shown in Fig.~\ref{fig:detection_zero_shot_dotav2}, the proposed method consistently demonstrates strong detection performance across all settings. In highly dense scenes, it successfully recovers most small objects under sparse point prompts, exhibiting accurate localization and robust category-level generalization. In contrast, SAM3, which primarily relies on high-level visual representations, struggles to reliably recall all instances in dense small-object scenarios on DOTA-v2.0.
Moreover, our method shows strong discriminative capability in suppressing cross-category false positives. When multiple object categories co-exist in the same scene, DEAL effectively filters out irrelevant objects and responds only to the prompted category, even under severe object density.
To further examine robustness under unseen data distributions, we evaluate our method on several Internet-sourced images with densely distributed small objects, including cartoon-style scenes and surveillance footage (Fig.~\ref{fig:detection_zero_shot_redbook}). Despite significant domain shifts and the absence of overlapping categories with the training data, our approach maintains strong detection performance and generalization.
Overall, these qualitative results demonstrate the effectiveness and robustness of the proposed point-prompt small object detection paradigm in both dense and unseen scenarios.
For more qualitative results across diverse domains, please refer to the Appendix.

\section{Discussions}
\label{sec:discussion}
In this study, we construct the large-scale TinySet-9M dataset and systematically benchmark existing label-efficient paradigms, revealing their severe performance degradation under reduced supervision in small object scenarios. Moreover, built upon the TinySet-9M, we propose the Point-Prompt Small Object Detection (P$^2$SOD) paradigm and the DEAL framework, which utilizes sparse inference-time point prompts to enable precise category-level localization. 
DEAL enables scalable and generalized small object detection without sacrificing localization accuracy.

To further advance the development of small object detection, this section focuses on answering the following three key questions.

\textit{-- Why point prompts instead of other forms of prompts?}

We adopt point prompts as they provide a minimal yet effective form of spatial interaction for small object detection, while serving as semantic anchors to enhance weak object representations.
First, a point is the simplest form of visual prompt, which makes it suitable for interactive settings.
Second, point prompts are particularly well-suited for small objects: due to their limited spatial extent, a point within the object can coarsely indicate the object itself.
Third, point prompts act as semantic anchors that strengthen feature association, alleviating the weak semantic cues of small objects. This is further consistent with their feature resolution, where small objects correspond to only a few feature points and richer prompts offer limited additional information.

\textit{-- How is P$^2$SOD fundamentally different from interactive segmentation or VLM-based prompting?}

P$^2$SOD shares a similar interaction form with prior prompt-based methods, but differs in how point prompts are interpreted and integrated into the detection pipeline. Specifically, P$^2$SOD treats point prompts as category-level semantic anchors that guide query generation and feature aggregation, rather than instance-specific cues for local refinement. In interactive segmentation (\textit{e.g.}, SAM), point prompts are used to isolate individual instances, whereas P$^2$SOD leverages them to activate all instances of the same category across the image. Compared to VLM-based prompting, which relies on high-level semantic alignment, P$^2$SOD directly conditions detection on spatial prompts, reducing ambiguity under weak visual cues.
This difference further leads to a fundamental advantage in interaction efficiency. In many small-object scenarios (\textit{e.g.}, remote sensing imagery), the number of instances $N$ is extremely large while the number of categories $C$ is small ($N \gg C$). Conventional instance-level interaction incurs an $O(N)$ annotation cost, which becomes impractical in dense scenes. In contrast, by elevating point prompts to category-level anchors, P$^2$SOD reduces the interaction complexity to $O(C)$. This shift from instance-wise to category-wise interaction significantly improves scalability, making the paradigm particularly suitable for dense small object detection.


\textit{-- What are the limitations of DEAL?}

The limitations of DEAL mainly lie in fine-grained category discrimination, applicability to large-object scenarios, and the reliance on task-specific thresholds during inference.
First, DEAL faces challenges when small object detection is coupled with fine-grained recognition tasks. The inherent lack of visual details in small objects causes severe feature overlap among similar categories, making single positive prompts insufficient for precise discrimination. To tackle this, future work could introduce negative point prompts as repulsive semantic anchors. By explicitly penalizing the activation of visually similar distractors, negative prompts can help carve out sharper decision boundaries in ambiguous scenes. Nevertheless, elegantly balancing this negative suppression without degrading the inherently weak visual responses of the target small objects poses a non-trivial architectural challenge, marking a promising direction for future exploration.
Second, DEAL is tailored for small and dense objects; its advantages diminish in large-object scenarios with clear semantics, where conventional fully supervised detectors remain more suitable. Future work could explore a cross-scale querying mechanism to enhance multi-scale detection capability, particularly for large objects.
Finally, the prompt-driven inference introduces several task-dependent thresholds, which, although empirically stable, add extra hyperparameters. Future work may explore adaptive alternatives to improve robustness and generality.

\section{Conclusion}
In this study, we expose a critical bottleneck in generalized small object detection: the inherent semantic deficiency of small objects undermines the effectiveness of existing detection paradigms. By systematically benchmarking these paradigms—facilitated by our introduced large-scale TinySet-9M dataset—we demonstrate that traditional supervision models suffer from severe localization and recognition collapse when confronted with information-starved visual signals. To transcend this representation challenge, we shift the perspective from purely training-time feature enhancement to inference-time semantic activation. We formalize the Point-Prompt Small Object Detection (P$^2$SOD) paradigm and the DEAL framework, establishing a novel interactive paradigm that utilizes sparse point prompts as category-level semantic anchors. By explicitly widening the activation disparity between small objects and background clutter, our method successfully compensates for vanishing object responses. This minimal-interaction mechanism not only achieves superior localization accuracy but also demonstrates robust generalization across unseen domains and novel categories.
While DEAL sets a strong foundation for generalized small object detection, resolving highly entangled semantic spaces—where small object detection is tightly coupled with fine-grained recognition—remains a formidable challenge. Future research could explore the incorporation of negative point prompts to enhance fine-grained recognition. 
Ultimately, this work defines a new trajectory for small object detection, encouraging a shift toward minimal-interaction paradigms as a principled solution to information-limited visual understanding.

\section*{Acknowledgments}
This research was supported by the National Key Research and Development Program under Grant 2024YFF1401003 and the National Natural Science Foundation of China under Grant 62271355.

\bibliographystyle{IEEEtran}
\bibliography{zhu}

@STRING{CVPR = "IEEE Conference on Computer Vision and Pattern Recognition"}

@STRING{ICCV = "IEEE International Conference on Computer Vision"}

@STRING{ECCV = "European Conference on Computer Vision"}

@STRING{AAAI = "AAAI Conference on Artificial Intelligence"}

@STRING{NIPS = "Advances in Neural Information Processing Systems"}

@STRING{WACV = "IEEE Workshops on Applications of Computer Vision"}

@STRING{ICPR = "International Conference on Pattern Recognition"}

@STRING{TPAMI = "IEEE Transactions on Pattern Analysis and Machine Intelligence"}

@STRING{IJCV = "International Journal of Computer Vision"}

@STRING{ISPRSJ = "ISPRS Journal of Photogrammetry and Remote Sensing"}

@article{aitodv2_2022_isprs,
    title={Detecting Tiny Objects in Aerial Images: A Normalized Wasserstein Distance and A New Benchmark},
    author={Xu, Chang and Wang, Jinwang and Yang, Wen and Yu, Huai and Yu, Lei and Xia, Gui-Song},
    journal={ISPRS Journal of Photogrammetry and Remote Sensing},
    volume={190},
    pages={79--93},
    year={2022},
}

@inproceedings{rfla_2022_eccv,
  title={RFLA: Gaussian Receptive Field based Label Assignment for Tiny Object Detection},
  author={Xu, Chang and Wang, Jinwang and Yang, Wen and Yu, Huai and Yu, Lei and Xia, Gui-Song},
  booktitle=ECCV,
  pages={526--543},
  year={2022},
  organization={Springer}
}

@article{DOTA2.0_2021_pami,
  title={Object detection in aerial images: A large-scale benchmark and challenges},
  author={Ding, Jian and Xue, Nan and Xia, Gui-Song and Bai, Xiang and Yang, Wen and Yang, Michael Ying and Belongie, Serge and Luo, Jiebo and Datcu, Mihai and Pelillo, Marcello and others},
  journal=TPAMI,
  volume={44},
  number={11},
  pages={7778--7796},
  year={2021},
  publisher={IEEE}
}

@article{soda_2023_pami,
  title={Towards large-scale small object detection: Survey and benchmarks},
  author={Cheng, Gong and Yuan, Xiang and Yao, Xiwen and Yan, Kebing and Zeng, Qinghua and Xie, Xingxing and Han, Junwei},
  journal={IEEE Transactions on Pattern Analysis and Machine Intelligence},
  year={2023},
  volume={45},
  number={11},
  pages={13467-13488},
  publisher={IEEE}
}

@article{visdrone_2021_pami,
  title={Detection and tracking meet drones challenge},
  author={Zhu, Pengfei and Wen, Longyin and Du, Dawei and Bian, Xiao and Fan, Heng and Hu, Qinghua and Ling, Haibin},
  journal={IEEE Transactions on Pattern Analysis and Machine Intelligence},
  volume={44},
  number={11},
  pages={7380--7399},
  year={2021},
  publisher={IEEE}
}

@article{PASCAL-VOC_2015-IJCV,
    title={The Pascal Visual Object Classes Challenge: A Retrospective},
    author={Everingham, Mark and Eslami, SM Ali and Van Gool, Luc and Williams, Christopher KI and Winn, John and Zisserman, Andrew},
    journal=IJCV,
    volume={111},
    number={1},
    pages={98--136},
    year={2015},
    publisher={IEEE}
}

@article{DIOR_2019_ISPRS,
    title={Object detection in optical remote sensing images: A survey and a new benchmark},
    author={Li, Ke and Wan, Gang and Cheng, Gong and Meng, Liqiu and Han, Junwei},
    journal=ISPRSJ,
    volume={159},
    pages={296--307},
    year={2020},
    publisher={Elsevier}
}

@article{FCOS_2022_TPAMI,
    title={Fcos: A simple and strong anchor-free object detector},
    author={Tian, Zhi and Shen, Chunhua and Chen, Hao and He, Tong},
    journal=TPAMI,
    year={2022},
    volume={44},
    number={4},
    pages={1922-1933},
    publisher={IEEE}
}

@inproceedings{COCO_2014_ECCV,
	title={Microsoft coco: Common objects in context},
	author={Lin, Tsung-Yi and Maire, Michael and Belongie, Serge and Hays, James and Perona, Pietro and Ramanan, Deva and Doll{\'a}r, Piotr and Zitnick, C Lawrence},
	booktitle=ECCV,
	pages={740--755},
	year={2014},
	organization={Springer}
}

@inproceedings{Faster-R-CNN_2015_NIPS,
    title = {{Faster R-CNN}: Towards Real-Time Object Detection with Region Proposal Networks},
    author = {Ren, Shaoqing and He, Kaiming and Girshick, Ross and Sun, Jian},
    booktitle = NIPS,
    pages = {91--99},
    year = {2015},
}

@inproceedings{Focal-Loss_2017_ICCV,
    author = {Lin, Tsung-Yi and Goyal, Priya and Girshick, Ross and He, Kaiming and Dollar, Piotr},
    title = {Focal Loss for Dense Object Detection},
    booktitle = ICCV,
    pages={2980--2988},
    year = {2017}
}

@InProceedings{FCOS_2019_ICCV,
    author = {Tian, Zhi and Shen, Chunhua and Chen, Hao and He, Tong},
    title = {{FCOS}: Fully Convolutional One-Stage Object Detection},
    booktitle = ICCV,
    pages={9627--9636},
    year = {2019}
}

@inproceedings{TinyPerson_2020_WACV,
    title={Scale match for tiny person detection},
    author={Yu, Xuehui and Gong, Yuqi and Jiang, Nan and Ye, Qixiang and Han, Zhenjun},
    booktitle=WACV,
    pages={1257--1265},
    year={2020}
}

@article{xview_2018_arXiv,
  title={xview: Objects in context in overhead imagery},
  author={Lam, Darius and Kuzma, Richard and McGee, Kevin and Dooley, Samuel and Laielli, Michael and Klaric, Matthew and Bulatov, Yaroslav and McCord, Brendan},
  journal={arXiv preprint arXiv:1802.07856},
  year={2018}
}

@inproceedings{rtdetr,
  title={Detrs beat yolos on real-time object detection},
  author={Zhao, Yian and Lv, Wenyu and Xu, Shangliang and Wei, Jinman and Wang, Guanzhong and Dang, Qingqing and Liu, Yi and Chen, Jie},
  booktitle={Proceedings of the IEEE/CVF conference on computer vision and pattern recognition},
  pages={16965--16974},
  year={2024}
}

@article{tod_ga,
  title={Label assignment matters: A gaussian assignment strategy for tiny object detection},
  author={Zhang, Feng and Zhou, Shilin and Wang, Yingqian and Wang, Xueying and Hou, Yi},
  journal={IEEE Transactions on Geoscience and Remote Sensing},
  year={2024},
  publisher={IEEE}
}

@INPROCEEDINGS{tod_simd,
  author={Shi, Shuohao and Fang, Qiang and Xu, Xin and Zhao, Tong},
  booktitle={2024 IEEE/RSJ International Conference on Intelligent Robots and Systems (IROS)}, 
  title={Similarity Distance-Based Label Assignment for Tiny Object Detection}, 
  year={2024},
  volume={},
  number={},
  pages={13711-13718},
  doi={10.1109/IROS58592.2024.10801448}}

@inproceedings{tod_hsfpn,
  title={HS-FPN: High frequency and spatial perception FPN for tiny object detection},
  author={Shi, Zican and Hu, Jing and Ren, Jie and Ye, Hengkang and Yuan, Xuyang and Ouyang, Yan and He, Jia and Ji, Bo and Guo, Junyu},
  booktitle={Proceedings of the AAAI Conference on Artificial Intelligence},
  volume={39},
  number={7},
  pages={6896--6904},
  year={2025}
}

@article{pointobbv2,
  title={Pointobb-v2: Towards simpler, faster, and stronger single point supervised oriented object detection},
  author={Ren, Botao and Yang, Xue and Yu, Yi and Luo, Junwei and Deng, Zhidong},
  journal={arXiv preprint arXiv:2410.08210},
  year={2024}
}

@article{pointobbv3,
  title={Pointobb-v3: Expanding performance boundaries of single point-supervised oriented object detection},
  author={Zhang, Peiyuan and Luo, Junwei and Yang, Xue and Yu, Yi and Li, Qingyun and Zhou, Yue and Jia, Xiaosong and Lu, Xudong and Chen, Jingdong and Li, Xiang and others},
  journal={International Journal of Computer Vision},
  pages={1--21},
  year={2025},
  publisher={Springer}
}

@inproceedings{p2bnet,
  title={Point-to-box network for accurate object detection via single point supervision},
  author={Chen, Pengfei and Yu, Xuehui and Han, Xumeng and Hassan, Najmul and Wang, Kai and Li, Jiachen and Zhao, Jian and Shi, Humphrey and Han, Zhenjun and Ye, Qixiang},
  booktitle={European Conference on Computer Vision},
  pages={51--67},
  year={2022},
  organization={Springer}
}

@article{point_teacher,
  title={Tiny object detection with single point supervision},
  author={Zhu, Haoran and Xu, Chang and Zhang, Ruixiang and Xu, Fang and Yang, Wen and Zhang, Haijian and Xia, Gui-Song},
  journal={ISPRS Journal of Photogrammetry and Remote Sensing},
  volume={227},
  pages={219--233},
  year={2025},
  publisher={Elsevier}
}

@inproceedings{pseco,
  title={Pseco: Pseudo labeling and consistency training for semi-supervised object detection},
  author={Li, Gang and Li, Xiang and Wang, Yujie and Wu, Yichao and Liang, Ding and Zhang, Shanshan},
  booktitle={European Conference on Computer Vision},
  pages={457--472},
  year={2022},
  organization={Springer}
}

@article{rexomni,
  title={Detect Anything via Next Point Prediction},
  author={Jiang, Qing and Huo, Junan and Chen, Xingyu and Xiong, Yuda and Zeng, Zhaoyang and Chen, Yihao and Ren, Tianhe and Yu, Junzhi and Zhang, Lei},
  journal={arXiv preprint arXiv:2510.12798},
  year={2025}
}

@inproceedings{groundingdino,
  title={Grounding dino: Marrying dino with grounded pre-training for open-set object detection},
  author={Liu, Shilong and Zeng, Zhaoyang and Ren, Tianhe and Li, Feng and Zhang, Hao and Yang, Jie and Jiang, Qing and Li, Chunyuan and Yang, Jianwei and Su, Hang and others},
  booktitle={European conference on computer vision},
  pages={38--55},
  year={2024},
  organization={Springer}
}

@article{qwen25vl,
  title={Qwen2. 5-vl technical report},
  author={Bai, Shuai and Chen, Keqin and Liu, Xuejing and Wang, Jialin and Ge, Wenbin and Song, Sibo and Dang, Kai and Wang, Peng and Wang, Shijie and Tang, Jun and others},
  journal={arXiv preprint arXiv:2502.13923},
  year={2025}
}

@article{labelnoise_dntod,
  title={Robust tiny object detection in aerial images amidst label noise},
  author={Zhu, Haoran and Xu, Chang and Yang, Wen and Zhang, Ruixiang and Zhang, Yan and Xia, Gui-Song},
  journal={arXiv preprint arXiv:2401.08056},
  year={2024}
}

@inproceedings{labelnoise_oamil,
  title={Robust object detection with inaccurate bounding boxes},
  author={Liu, Chengxin and Wang, Kewei and Lu, Hao and Cao, Zhiguo and Zhang, Ziming},
  booktitle={European Conference on Computer Vision},
  pages={53--69},
  year={2022},
  organization={Springer}
}

@inproceedings{labelnoise_ssddet,
  title={Spatial self-distillation for object detection with inaccurate bounding boxes},
  author={Wu, Di and Chen, Pengfei and Yu, Xuehui and Li, Guorong and Han, Zhenjun and Jiao, Jianbin},
  booktitle={Proceedings of the IEEE/CVF International Conference on Computer Vision},
  pages={6855--6865},
  year={2023}
}

@inproceedings{semi_softteacher,
  title={End-to-end semi-supervised object detection with soft teacher},
  author={Xu, Mengde and Zhang, Zheng and Hu, Han and Wang, Jianfeng and Wang, Lijuan and Wei, Fangyun and Bai, Xiang and Liu, Zicheng},
  booktitle={Proceedings of the IEEE/CVF international conference on computer vision},
  pages={3060--3069},
  year={2021}
}

@inproceedings{sparse_comining,
  title={Co-mining: Self-supervised learning for sparsely annotated object detection},
  author={Wang, Tiancai and Yang, Tong and Cao, Jiale and Zhang, Xiangyu},
  booktitle={Proceedings of the AAAI conference on artificial intelligence},
  volume={35},
  number={4},
  pages={2800--2808},
  year={2021}
}

@inproceedings{sparse_sparsedet,
  title={Sparsedet: Improving sparsely annotated object detection with pseudo-positive mining},
  author={Suri, Saksham and Rambhatla, Saketh and Chellappa, Rama and Shrivastava, Abhinav},
  booktitle={Proceedings of the IEEE/CVF International Conference on Computer Vision},
  pages={6770--6781},
  year={2023}
}

@article{point2rboxv3,
  title={Point2RBox-v3: Self-Bootstrapping from Point Annotations via Integrated Pseudo-Label Refinement and Utilization},
  author={Zhang, Teng and Fan, Ziqian and Liu, Mingxin and Zhang, Xin and Lu, Xudong and Li, Wentong and Zhou, Yue and Yu, Yi and Li, Xiang and Yan, Junchi and others},
  journal={arXiv preprint arXiv:2509.26281},
  year={2025}
}

@inproceedings{sparse_calibratedteacher,
  title={Calibrated teacher for sparsely annotated object detection},
  author={Wang, Haohan and Liu, Liang and Zhang, Boshen and Zhang, Jiangning and Zhang, Wuhao and Gan, Zhenye and Wang, Yabiao and Wang, Chengjie and Wang, Haoqian},
  booktitle={Proceedings of the AAAI Conference on Artificial Intelligence},
  volume={37},
  number={2},
  pages={2519--2527},
  year={2023}
}

@article{sam3,
  title={Sam 3: Segment anything with concepts},
  author={Carion, Nicolas and Gustafson, Laura and Hu, Yuan-Ting and Debnath, Shoubhik and Hu, Ronghang and Suris, Didac and Ryali, Chaitanya and Alwala, Kalyan Vasudev and Khedr, Haitham and Huang, Andrew and others},
  journal={arXiv preprint arXiv:2511.16719},
  year={2025}
}

@ARTICLE{DCFL++,
  author={Xu, Chang and Zhang, Ruixiang and Yang, Wen and Zhu, Haoran and Xu, Fang and Ding, Jian and Xia, Gui-Song},
  journal={IEEE Transactions on Pattern Analysis and Machine Intelligence}, 
  title={Oriented Tiny Object Detection: A Dataset, Benchmark, and Dynamic Unbiased Learning}, 
  year={2025},
  volume={},
  number={},
  pages={1-18},
  doi={10.1109/TPAMI.2025.3634161}}

@article{groundedsam,
  title={Grounded sam: Assembling open-world models for diverse visual tasks},
  author={Ren, Tianhe and Liu, Shilong and Zeng, Ailing and Lin, Jing and Li, Kunchang and Cao, He and Chen, Jiayu and Huang, Xinyu and Chen, Yukang and Yan, Feng and others},
  journal={arXiv preprint arXiv:2401.14159},
  year={2024}
}

@article{sardet100k,
  title={Sardet-100k: Towards open-source benchmark and toolkit for large-scale sar object detection},
  author={Li, Yuxuan and Li, Xiang and Li, Weijie and Hou, Qibin and Liu, Li and Cheng, Ming-Ming and Yang, Jian},
  journal={Advances in Neural Information Processing Systems},
  volume={37},
  pages={128430--128461},
  year={2024}
}

@article{lsssdd,
  title={LS-SSDD-v1. 0: A deep learning dataset dedicated to small ship detection from large-scale Sentinel-1 SAR images},
  author={Zhang, Tianwen and Zhang, Xiaoling and Ke, Xiao and Zhan, Xu and Shi, Jun and Wei, Shunjun and Pan, Dece and Li, Jianwei and Su, Hao and Zhou, Yue and others},
  journal={Remote Sensing},
  volume={12},
  number={18},
  pages={2997},
  year={2020},
  publisher={MDPI}
}

@article{hrsid,
  title={HRSID: A high-resolution SAR images dataset for ship detection and instance segmentation},
  author={Wei, Shunjun and Zeng, Xiangfeng and Qu, Qizhe and Wang, Mou and Su, Hao and Shi, Jun},
  journal={Ieee Access},
  volume={8},
  pages={120234--120254},
  year={2020},
  publisher={IEEE}
}

@article{sirst,
  title={SIRST-5K: Exploring massive negatives synthesis with self-supervised learning for robust infrared small target detection},
  author={Lu, Yahao and Lin, Yupei and Wu, Han and Xian, Xiaoyu and Shi, Yukai and Lin, Liang},
  journal={IEEE Transactions on Geoscience and Remote Sensing},
  volume={62},
  pages={1--11},
  year={2024},
  publisher={IEEE}
}

@inproceedings{irstd-1k,
  title={ISNet: Shape matters for infrared small target detection},
  author={Zhang, Mingjin and Zhang, Rui and Yang, Yuxiang and Bai, Haichen and Zhang, Jing and Guo, Jie},
  booktitle={Proceedings of the IEEE/CVF conference on computer vision and pattern recognition},
  pages={877--886},
  year={2022}
}

@article{RGBTDronePerson,
  title={Drone-based RGBT tiny person detection},
  author={Zhang, Yan and Xu, Chang and Yang, Wen and He, Guangjun and Yu, Huai and Yu, Lei and Xia, Gui-Song},
  journal={ISPRS Journal of Photogrammetry and Remote Sensing},
  volume={204},
  pages={61--76},
  year={2023},
  publisher={Elsevier}
}

@article{rgbt-tiny,
  title={Visible-thermal tiny object detection: A benchmark dataset and baselines},
  author={Ying, Xinyi and Xiao, Chao and An, Wei and Li, Ruojing and He, Xu and Li, Boyang and Cao, Xu and Li, Zhaoxu and Wang, Yingqian and Hu, Mingyuan and others},
  journal={IEEE Transactions on Pattern Analysis and Machine Intelligence},
  year={2025},
  publisher={IEEE}
}

@inproceedings{sku110k,
  title={Precise detection in densely packed scenes},
  author={Goldman, Eran and Herzig, Roei and Eisenschtat, Aviv and Goldberger, Jacob and Hassner, Tal},
  booktitle={Proceedings of the IEEE/CVF conference on computer vision and pattern recognition},
  pages={5227--5236},
  year={2019}
}

@article{jhu_crowd++,
  title={Jhu-crowd++: Large-scale crowd counting dataset and a benchmark method},
  author={Sindagi, Vishwanath A and Yasarla, Rajeev and Patel, Vishal M},
  journal={IEEE transactions on pattern analysis and machine intelligence},
  volume={44},
  number={5},
  pages={2594--2609},
  year={2020},
  publisher={IEEE}
}

@article{crowdhuman,
  title={Crowdhuman: A benchmark for detecting human in a crowd},
  author={Shao, Shuai and Zhao, Zijian and Li, Boxun and Xiao, Tete and Yu, Gang and Zhang, Xiangyu and Sun, Jian},
  journal={arXiv preprint arXiv:1805.00123},
  year={2018}
}

@article{livecell,
  title={LIVECell—A large-scale dataset for label-free live cell segmentation},
  author={Edlund, Christoffer and Jackson, Timothy R and Khalid, Nabeel and Bevan, Nicola and Dale, Timothy and Dengel, Andreas and Ahmed, Sheraz and Trygg, Johan and Sj{\"o}gren, Rickard},
  journal={Nature methods},
  volume={18},
  number={9},
  pages={1038--1045},
  year={2021},
  publisher={Nature Publishing Group US New York}
}

@inproceedings{tyc,
  title={The TYC Dataset for Understanding Instance-Level Semantics and Motions of Cells in Microstructures},
  author={Reich, Christoph and Prangemeier, Tim and Koeppl, Heinz},
  booktitle={Proceedings of the IEEE/CVF International Conference on Computer Vision},
  pages={3940--3951},
  year={2023}
}

@article{visalgae,
  title={VisAlgae 2023: A Dataset and Challenge for Algae Detection in Microscopy Images},
  author={Sun, Mingxuan and Jiang, Juntao and Yang, Zhiqiang and Kong, Shenao and Qi, Jiamin and Shang, Jianru and Luo, Shuangling and Sun, Wanfa and Wang, Tianyi and Wang, Yanqi and others},
  journal={arXiv preprint arXiv:2505.20687},
  year={2025}
}

@article{pest24,
  title={Pest24: A large-scale very small object data set of agricultural pests for multi-target detection},
  author={Wang, Qi-Jin and Zhang, Sheng-Yu and Dong, Shi-Feng and Zhang, Guang-Cai and Yang, Jin and Li, Rui and Wang, Hong-Qiang},
  journal={Computers and electronics in agriculture},
  volume={175},
  pages={105585},
  year={2020},
  publisher={Elsevier}
}

@article{tacna_olive,
  title={The Detection and Counting of Olive Tree Fruits Using Deep Learning Models in Tacna, Per{\'u}},
  author={Osco-Mamani, Erbert and Santana-Carbajal, Oliver and Chaparro-Cruz, Israel and Ochoa-Donoso, Daniel and Alcazar-Alay, Sylvia},
  journal={AI},
  volume={6},
  number={2},
  pages={25},
  year={2025},
  publisher={MDPI}
}

@inproceedings{related_SimD_strategy,
  title={Similarity distance-based label assignment for tiny object detection},
  author={Shi, Shuohao and Fang, Qiang and Xu, Xin and Zhao, Tong},
  booktitle={2024 IEEE/RSJ International Conference on Intelligent Robots and Systems (IROS)},
  pages={13711--13718},
  year={2024},
  organization={IEEE}
}

@ARTICLE{related_FRLI_Net,
  author={Chen, Penglei and Wang, Jiangtao and Zhang, Zhiwei and He, Cheng},
  journal={IEEE Signal Processing Letters}, 
  title={FRLI-Net: Feature Reconstruction and Learning Interaction Network for Tiny Object Detection in Remote Sensing Images}, 
  year={2025},
  volume={32},
  number={},
  pages={2159-2163},
  doi={10.1109/LSP.2025.3546859}}

@InProceedings{related_SET,
    author    = {Sun, Huixin and Wang, Runqi and Li, Yanjing and Yang, Linlin and Lin, Shaohui and Cao, Xianbin and Zhang, Baochang},
    title     = {SET: Spectral Enhancement for Tiny Object Detection},
    booktitle = {Proceedings of the IEEE/CVF Conference on Computer Vision and Pattern Recognition (CVPR)},
    month     = {June},
    year      = {2025},
    pages     = {4713-4723}
}

@InProceedings{related_PFIM,
    author    = {Bian, Jinghao and Feng, Mingtao and Dong, Weisheng and Wu, Fangfang and Luo, Jianqiao and Wang, Yaonan and Shi, Guangming},
    title     = {Feature Information Driven Position Gaussian Distribution Estimation for Tiny Object Detection},
    booktitle = {Proceedings of the IEEE/CVF Conference on Computer Vision and Pattern Recognition (CVPR)},
    month     = {June},
    year      = {2025},
    pages     = {30376-30386}
}

@inproceedings{semi_consistent_teacher,
  title={Consistent-teacher: Towards reducing inconsistent pseudo-targets in semi-supervised object detection},
  author={Wang, Xinjiang and Yang, Xingyi and Zhang, Shilong and Li, Yijiang and Feng, Litong and Fang, Shijie and Lyu, Chengqi and Chen, Kai and Zhang, Wayne},
  booktitle={Proceedings of the IEEE/CVF conference on computer vision and pattern recognition},
  pages={3240--3249},
  year={2023}
}

@article{ALOD,
  title={Minimizing Sample Redundancy for Label-efficient Object Detection in Aerial Images},
  author={Zhang, Ruixiang and Xu, Chang and Zhu, Haorao and Xu, Fang and Yang, Wen and Zhang, Haijian and Xia, Gui-Song},
  journal={IEEE Transactions on Geoscience and Remote Sensing},
  year={2025},
  publisher={IEEE}
}

@inproceedings{VLM_CLIP,
  title={Learning transferable visual models from natural language supervision},
  author={Radford, Alec and Kim, Jong Wook and Hallacy, Chris and Ramesh, Aditya and Goh, Gabriel and Agarwal, Sandhini and Sastry, Girish and Askell, Amanda and Mishkin, Pamela and Clark, Jack and others},
  booktitle={International conference on machine learning},
  pages={8748--8763},
  year={2021},
  organization={PmLR}
}

@inproceedings{GASSL,
  title={Geography-aware self-supervised learning},
  author={Ayush, Kumar and Uzkent, Burak and Meng, Chenlin and Tanmay, Kumar and Burke, Marshall and Lobell, David and Ermon, Stefano},
  booktitle={Proceedings of the IEEE/CVF International Conference on Computer Vision},
  pages={10181--10190},
  year={2021}
}

@inproceedings{CACO,
  title={Change-aware sampling and contrastive learning for satellite images},
  author={Mall, Utkarsh and Hariharan, Bharath and Bala, Kavita},
  booktitle={Proceedings of the IEEE/CVF Conference on Computer Vision and Pattern Recognition},
  pages={5261--5270},
  year={2023}
}

@article{TOV,
  title={TOV: The original vision model for optical remote sensing image understanding via self-supervised learning},
  author={Tao, Chao and Qi, Ji and Zhang, Guo and Zhu, Qing and Lu, Weipeng and Li, Haifeng},
  journal={IEEE Journal of Selected Topics in Applied Earth Observations and Remote Sensing},
  volume={16},
  pages={4916--4930},
  year={2023},
  publisher={IEEE}
}

@inproceedings{Scale-MAE,
  title={Scale-mae: A scale-aware masked autoencoder for multiscale geospatial representation learning},
  author={Reed, Colorado J and Gupta, Ritwik and Li, Shufan and Brockman, Sarah and Funk, Christopher and Clipp, Brian and Keutzer, Kurt and Candido, Salvatore and Uyttendaele, Matt and Darrell, Trevor},
  booktitle={Proceedings of the IEEE/CVF International Conference on Computer Vision},
  pages={4088--4099},
  year={2023}
}

@inproceedings{SatLas,
  title={Satlaspretrain: A large-scale dataset for remote sensing image understanding},
  author={Bastani, Favyen and Wolters, Piper and Gupta, Ritwik and Ferdinando, Joe and Kembhavi, Aniruddha},
  booktitle={Proceedings of the IEEE/CVF International Conference on Computer Vision},
  pages={16772--16782},
  year={2023}
}

@article{RingMo,
  title={RingMo: A remote sensing foundation model with masked image modeling},
  author={Sun, Xian and Wang, Peijin and Lu, Wanxuan and Zhu, Zicong and Lu, Xiaonan and He, Qibin and Li, Junxi and Rong, Xuee and Yang, Zhujun and Chang, Hao and others},
  journal={IEEE Transactions on Geoscience and Remote Sensing},
  volume={61},
  pages={1--22},
  year={2022},
  publisher={IEEE}
}

@inproceedings{SkySense,
  title={Skysense: A multi-modal remote sensing foundation model towards universal interpretation for earth observation imagery},
  author={Guo, Xin and Lao, Jiangwei and Dang, Bo and Zhang, Yingying and Yu, Lei and Ru, Lixiang and Zhong, Liheng and Huang, Ziyuan and Wu, Kang and Hu, Dingxiang and others},
  booktitle={Proceedings of the IEEE/CVF Conference on Computer Vision and Pattern Recognition},
  pages={27672--27683},
  year={2024}
}

@article{MTP,
  title={MTP: Advancing remote sensing foundation model via multitask pretraining},
  author={Wang, Di and Zhang, Jing and Xu, Minqiang and Liu, Lin and Wang, Dongsheng and Gao, Erzhong and Han, Chengxi and Guo, Haonan and Du, Bo and Tao, Dacheng and others},
  journal={IEEE Journal of Selected Topics in Applied Earth Observations and Remote Sensing},
  volume={17},
  pages={11632--11654},
  year={2024},
  publisher={IEEE}
}

@inproceedings{backbone_resnet,
  title={Deep residual learning for image recognition},
  author={He, Kaiming and Zhang, Xiangyu and Ren, Shaoqing and Sun, Jian},
  booktitle={Proceedings of the IEEE conference on computer vision and pattern recognition},
  pages={770--778},
  year={2016}
}

@article{backbone_pvtv2,
  title={Pvt v2: Improved baselines with pyramid vision transformer},
  author={Wang, Wenhai and Xie, Enze and Li, Xiang and Fan, Deng-Ping and Song, Kaitao and Liang, Ding and Lu, Tong and Luo, Ping and Shao, Ling},
  journal={Computational visual media},
  volume={8},
  number={3},
  pages={415--424},
  year={2022},
  publisher={TUP}
}

@inproceedings{backbone_swimt,
  title={Swin transformer: Hierarchical vision transformer using shifted windows},
  author={Liu, Ze and Lin, Yutong and Cao, Yue and Hu, Han and Wei, Yixuan and Zhang, Zheng and Lin, Stephen and Guo, Baining},
  booktitle={Proceedings of the IEEE/CVF international conference on computer vision},
  pages={10012--10022},
  year={2021}
}

@article{backbone_dinov3,
  title={Dinov3},
  author={Sim{\'e}oni, Oriane and Vo, Huy V and Seitzer, Maximilian and Baldassarre, Federico and Oquab, Maxime and Jose, Cijo and Khalidov, Vasil and Szafraniec, Marc and Yi, Seungeun and Ramamonjisoa, Micha{\"e}l and others},
  journal={arXiv preprint arXiv:2508.10104},
  year={2025}
}

@misc{DFL-BundesligaDataset,
    author = {enddl22},
    title = {20k bounding boxes, DFLBundesliga Data Shootout},
    year = {2021},
    howpublished = {\url{https://www.kaggle.com/datasets/enddl22/bounding-boxes-dflbundesliga-data-shootout}},
    note = {Kaggle}
}

@article{SH17Dataset,
title = {SH17: A dataset for human safety and personal protective equipment detection in manufacturing industry},
journal = {Journal of Safety Science and Resilience},
volume = {6},
number = {2},
pages = {175-185},
year = {2025},
issn = {2666-4496},
doi = {https://doi.org/10.1016/j.jnlssr.2024.09.002},
url = {https://www.sciencedirect.com/science/article/pii/S266644962400077X},
author = {Hafiz Mughees Ahmad and Afshin Rahimi},
}

@INPROCEEDINGS{SCUTHEADDataset,
  author={Peng, Dezhi and Sun, Zikai and Chen, Zirong and Cai, Zirui and Xie, Lele and Jin, Lianwen},
  booktitle={2018 24th International Conference on Pattern Recognition (ICPR)}, 
  title={Detecting Heads using Feature Refine Net and Cascaded Multi-scale Architecture}, 
  year={2018},
  volume={},
  number={},
  pages={2528-2533},
  doi={10.1109/ICPR.2018.8545068}}

@misc{IRwildlifeDataset,
    author = {wangsheng0352},
    title = {Infrared Small Object Detection for Wildlife Conservation},
    year = {2024},
    howpublished = {\url{https://www.kaggle.com/datasets/wangsheng0352/ir-wildlife}},
    note = {Kaggle}
}

@misc{LicensePlateDetectionDataset,
    author = {Fares Elmenshawii},
    title = {License Plate Detection Dataset},
    year = {2024},
    howpublished = {\url{https://www.kaggle.com/datasets/fareselmenshawii/license-plate-dataset}},
    note = {Kaggle}
}

@inproceedings{irtdetr,
  title={Interactive object detection for tiny objects in large remotely sensed images},
  author={Burges, Marvin and Zambanini, Sebastian and Sablatnig, Robert},
  booktitle={2025 IEEE/CVF Winter Conference on Applications of Computer Vision (WACV)},
  pages={4704--4713},
  year={2025},
  organization={IEEE}
}

\vfill

\end{document}